\documentclass[10pt,twocolumn,letterpaper]{article}

\usepackage{cvpr}              

\definecolor{cvprblue}{rgb}{0.21,0.49,0.74}
\usepackage[pagebackref,breaklinks,colorlinks,allcolors=cvprblue]{hyperref}
\usepackage{booktabs}
\usepackage{kotex}
\usepackage{multicol}
\usepackage{multirow}
\usepackage{amsmath}
\usepackage{verbatim}
\usepackage{algorithm}
\usepackage{algpseudocode}
\usepackage{algorithmicx}
\usepackage{algcompatible}

\def\paperID{12398} 
\def\confName{CVPR}
\def\confYear{2025}

\title{Two is Better than One: \\Efficient Ensemble Defense for Robust and Compact Models}

\author{Yoojin Jung \quad Byung Cheol Song\\
Department of Electrical and Computer Engineering, Inha University\\
Incheon, Republic of Korea\\
{\tt\small yooon0505@gmail.com, }
{\tt\small bcsong@inha.ac.kr}
}

\begin{document}
\maketitle
\begin{abstract}
Deep learning-based computer vision systems adopt complex and large architectures to improve performance, yet they face challenges in deployment on resource-constrained mobile and edge devices. To address this issue, model compression techniques such as pruning, quantization, and matrix factorization have been proposed; however, these compressed models are often highly vulnerable to adversarial attacks. We introduce the \textbf{Efficient Ensemble Defense (EED)} technique, which diversifies the compression of a single base model based on different pruning importance scores and enhances ensemble diversity to achieve high adversarial robustness and resource efficiency. EED dynamically determines the number of necessary sub-models during the inference stage, minimizing unnecessary computations while maintaining high robustness. On the CIFAR-10 and SVHN datasets, EED demonstrated state-of-the-art robustness performance compared to existing adversarial pruning techniques, along with an inference speed improvement of up to 1.86 times. This proves that EED is a powerful defense solution in resource-constrained environments.
\end{abstract}    
\section{Introduction}
\label{sec:intro}

Convolutional Neural Networks (CNNs) have shown remarkable performance across various computer vision tasks, such as image classification, object detection, and semantic segmentation\cite{krizhevsky2009learning, redmon2016you, simonyan2014two}. However, CNNs have become increasingly wider and deeper to achieve higher performance, leading to complex architectures with vast numbers of parameters. This complexity inherently results in significant computational costs and storage demands, making deployment on resource-constrained edge devices challenging\cite{wang2020dual,chen2019collaborative}. To address this challenge, a number of model compression techniques, including pruning\cite{han2015learning, li2016pruning}, quantization\cite{chen2015compressing}, and matrix decomposition\cite{denil2013predicting}, have been proposed to minimize model size and computational load without compromising accuracy.

\begin{figure}[t]
  \centering
  \includegraphics[width=0.9\linewidth]{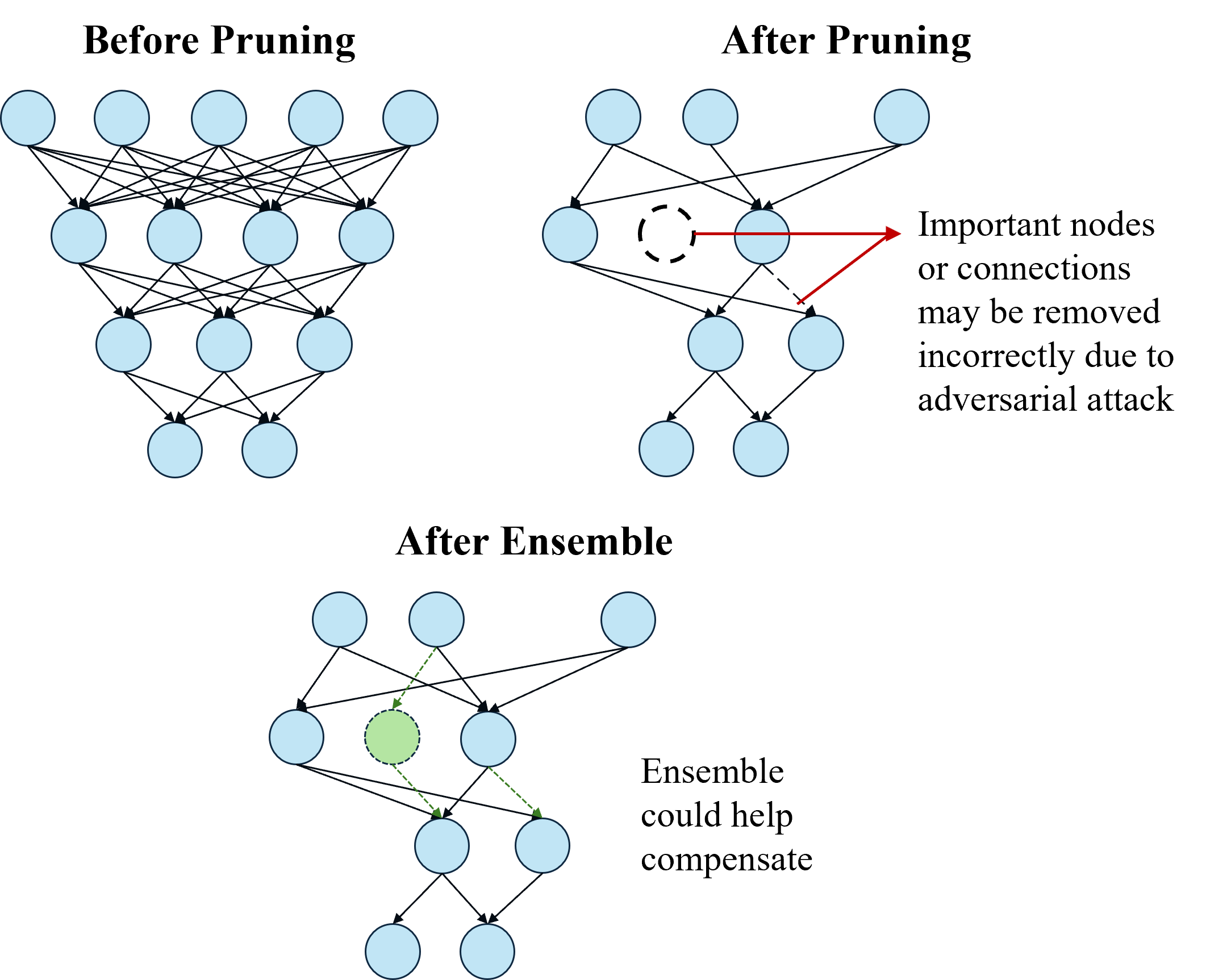}
  \caption{Challenges in AP and the potential of ensemble methods: AP weakens the network by removing critical parameters, reducing robustness and increasing vulnerability to adversarial attacks. An ensemble approach mitigates pruning effects by combining pruned models, enhancing robustness through their collective strength (illustrated in green).}
  \label{fig:1}
  \vspace{-1mm}
\end{figure}

Meanwhile, deep learning models are highly vulnerable to adversarial attacks that induce false predictions by disturbing the input image with small perturbations\cite{szegedy2014intriguing, goodfellow2015explaining}. This vulnerability is particularly pronounced in sparsely compressed models, which are even more sensitive to adversarial attacks than standard models\cite{shumailov2019compress, wu2021wider}. Consequently, common adversarial defense techniques such as adversarial training (AT) \cite{kurakin2016adversarial, madry2018PGD}, label smoothing \cite{zhang2019theoretically,wang2019improving}, and defensive distillation\cite{papernot2016distillation}, while generally effective, often experience a degradation in performance when applied to compressed models. 
This indicates that existing compression techniques may compromise adversarial robustness even if they maintain accuracy on clean test sets. In response, adversarial pruning (AP), a model compression technique that considers adversarial robustness\cite{sehwag2020hydra, ye2019adversarial, zhao2023holistic}, has emerged; however, AP still exhibits lower robustness compared to non-compressed defense models\cite{piras2024adversarial}.

We assume that this degradation in robustness arises from reduced generalization capability and information loss caused by pruning. Specifically, adversarial attacks may cause the model to misidentify and even remove critical nodes or connections during the pruning process. In Sec.\ref{sec:3.3}, we validate this assumption by observing importance scores, which reflect how each component contributes to the loss function through the gradients of the weights\cite{yu2018nisp}. Our findings demonstrate that adversarial examples significantly impact the importance scores of individual parameters.

Ensemble defense\cite{tramer2017ensemble, deng2024understanding} has recently gained attention in adversarial defense research for enhancing the stability and robustness of multiple base classifiers against attacks. As illustrated in Fig.\ref{fig:1}, we believe that ensemble defense methods can help mitigate generalization degradation and information loss during the pruning process. Since ensemble learning can compensate for deficiencies in individual compressed models due to removed parameters by other models and offset individual errors by combining outputs. However, traditional ensemble defenses inherently suffer from a large capacity, as they require multiple base models\cite{he2017adversarial, pang2019improving}. This raises an important question: \textit{Is it possible to develop an ensemble defense algorithm that not only achieves robustness but also results in a compact compressed model?}

As an answer to this, we propose Efficient Ensemble Defense (EED). Instead of using multiple base models, we compress a single base model by employing different pruning metrics and then utilize these compressed sub-models in the ensemble. This approach reduces the overall capacity compared to traditional ensemble defense models that rely on multiple base models. Given that using smaller models in an ensemble can lead to reduced defensive efficacy\cite{wu2021boosting}, we enhance the variety of the compressed models by incorporating a diversity term in the ensemble selection process, thereby improving ensemble effectiveness while maintaining robustness. Additionally, by adopting inference-efficient ensemble\cite{li2023towards}, we dynamically add sub-networks and perform ensemble operations as needed during inference, improving overall efficiency.

The main contributions of our work are as follows:
\begin{itemize}
    \item[$\bullet$] 
    Propose EED, that overcomes the limitations of traditional AP by enhancing the ensemble diversity and compensating for individual errors, thereby achieving robustness against adversarial attacks while being compact.
        
    \item[$\bullet$] 
    Reduce memory by using a single base model in ensemble, and decrease computational costs by dynamically adding models only when necessary during inference.
        
    \item[$\bullet$] 
    Achieve state-of-the-art performance across various attack benchmarks on the CIFAR-10 and SVHN datasets.
\end{itemize}

\section{Related Works}
\label{sec:related works}
\subsection{Adversarial Attack and Defense}
Adversarial attacks exploit the inherent vulnerabilities of deep learning models by adding specific noise to the input, thereby inducing incorrect predictions. These specially crafted inputs, designed to deceive deep learning models, are called adversarial examples. Early studies highlighted the susceptibility of deep learning models to adversarial examples\cite{szegedy2014intriguing}. Then, more potent attacks such as FGSM\cite{goodfellow2015explaining}, PGD\cite{madry2018PGD}, and AutoAttack\cite{croce2020reliable} have emerged. 

On the other hand, defense methods against attacks have also been evolved. The most well-known adversarial training (AT)\cite{kurakin2016adversarial, madry2018PGD, carlini2017towards} enhanced robustness by training adversarial examples. Other strategies involve regularizing geometric properties such as the curvature of the loss function~\cite{wang2019improving, wu2020adversarial}, or restoring potentially attacked images to purified versions\cite{nie2022diffusion, yoon2021adversarial}. Adversarial pruning (AP) and ensemble defense are also one of adversarial defenses (details are given in Sec.\ref{sec:3.1}.)

\subsection{Adversarial Pruning}
AP is a pruning technique (primarily weight pruning) that considers the adversarial context, aiming to reduce the complexity of neural networks while preserving robustness against adversarial attacks. HYDRA demonstrated that empirical adversarial robustness can be achieved through a lowest weight magnitude (LWM) pruning method. R-ADMM\cite{ye2019adversarial} and HARP\cite{zhao2023holistic} incorporated the alternating direction method of multipliers (ADMM) framework into LWM-based pruning to enhance robustness. In some studies~\cite{guo2018sparse, liao2022achieving, li2020towards}, a traditional pruning method, i.e., the Lottery Ticket Hypothesis, was integrated with AT.
However, they all failed to maintain generalized robustness at higher sparsity rates ($s_r$)\cite{han2015deep, kundu2021dnr, timpl2022understanding}, and struggled to defend against adaptive attacks\cite{dhillon2018stochastic, jian2022pruning}. This suggests that conventional AP has not fully resolved the trade-off between model capacity and robustness.

\subsection{Ensemble Defense}
Ensemble defense (ED), which applies ensemble learning to adversarial defense, has been actively studied due to its promising performance\cite{deng2024understanding,tramer2017ensemble}. The core of ED is to improve predictions through an ensemble diversity loss that considers adversarial robustness during training. ADP\cite{pang2019improving} used Shannon entropy and geometric diversity for uncertainty, while GAL\cite{kariyappa2019improving} utilized cosine similarities between model loss gradients to formalize diversity. DVERGE\cite{yang2020dverge} introduced vulnerability diversity to ensure that each model is robust to the weaknesses of others, and MOL\cite{cheng2022more} increased perturbation diversity by using expert models trained on different attacks.
Thus, ED has significantly enhanced robustness by reducing the risk of overfitting to adversarial examples and diversifying the decision boundary, though with increased model capacity and computational costs.

\section{Preliminaries}

\subsection{Notations}\label{sec:3.1}

\noindent
\textbf{Adversarial Attack} 
Let \( x \in \mathbb{R}^d \) be an input example, \( y \in \{1, \ldots, C\} \) the true label, and \( f(\cdot) \) a classifier with parameters \( \theta \). The classifier’s prediction for input \( x \) is given by \( f(x) = \arg \max_{i \in \{1, \ldots, C\}} f_i(x) \), where \( f_i(x) \) denotes the predicted probability for class \( i \). An adversarial example \( x' \) is crafted by adding a perturbation \( \delta \) to \( x \), aiming to maximize the model's prediction error while keeping \( x' \) close to \( x \) under a norm constraint. Formally, an adversarial attack is defined by:
\begin{equation}
    x' = x + \delta\quad \\= \arg\max_{\|x-x'\| < \epsilon} \mathcal{L}(f(x'), y)
\end{equation}
where \( \mathcal{L} \) is the loss function and \( \epsilon \) is the perturbation budget that constrains the strength of the attack under an \( \ell_p \)-norm.

\noindent
\textbf{Adversarial Training}
AT is a defense method that improves model robustness by training on adversarially perturbed samples alongside clean samples. The objective for AT is usually formulated by:
\begin{equation}
    \min_{\theta} \mathbb{E}_{(x, y) \sim \mathcal{D}} \left[ \max_{\delta: \|\delta\|_p \leq \epsilon} \mathcal{L}(f(x'; \theta), y) \right]
\end{equation}
where \( \mathcal{D} \) is the data distribution.

\noindent
\textbf{Adversarial Pruning}
Given a pre-trained model $f(\theta)$ under AT, we aim to maintain robustness while applying a pruning mask \( M \) across layers to achieve compression.
The general pruning is carried out by removing parameters until the network reaches a desired $s_r$. So, pruning involves creating \( M \), formalized as follows:
\begin{equation}
    M = \arg\min_{\|\mathbf{m}\|_0 \leq 1-s_r} \, \mathcal{L}(f(\theta) \otimes m, x, y).
\end{equation}
\begin{equation}
    m = 
    \begin{cases}
        0, & \text{if } \theta_i  \text{ is to be pruned}\\
        1, & \text{otherwise}
    \end{cases}
    , \quad \forall \, \theta_i \in \theta
\end{equation} 
For AP, we follow the strategy from HARP\cite{zhao2023holistic}.
To enhance adversarial robustness, inner maximization generates adversarial examples, targeting a worst-case loss under the PGD attacks:
\begin{equation}
   \max_{\delta} \mathcal{L}_{r}(f(\theta) \odot M, x + \delta, y)
\end{equation}
where \( \mathcal{L}_{r} \) defines the robust loss function, incorporating attack-specific parameters.
And the global non-uniform pruning strategy jointly optimizes a compression rate \( c_r \) and importance scores \( I \) across layers. This is defined by:
\begin{equation}
    \begin{split}
    \min_{c_r, I} \mathbb{E}_{(x,y) \sim D} \left[ \max_{\delta} \mathcal{L}_{r}(f(\theta) \odot M, x + \delta, y) \right] \\+ \varphi \cdot \mathcal{L}_{p}(f(\theta) \odot M, a_t)
    \end{split}
    \label{eq:pruning}
\end{equation}
where \( \mathcal{L}_{p} \) enforces target compression \( a_t \) across all layers \( L \) to guide non-uniform pruning rates. 

\subsection{Pruning Importance Score}\label{sec:3.2}
The importance score's definition varies by pruning method. This section details its definition and computation, as discussed in Sec.\ref{sec:3.1}.

\begin{figure*}[t]
  \centering
  \begin{subfigure}{0.35\linewidth}
    \includegraphics[width=1\linewidth]{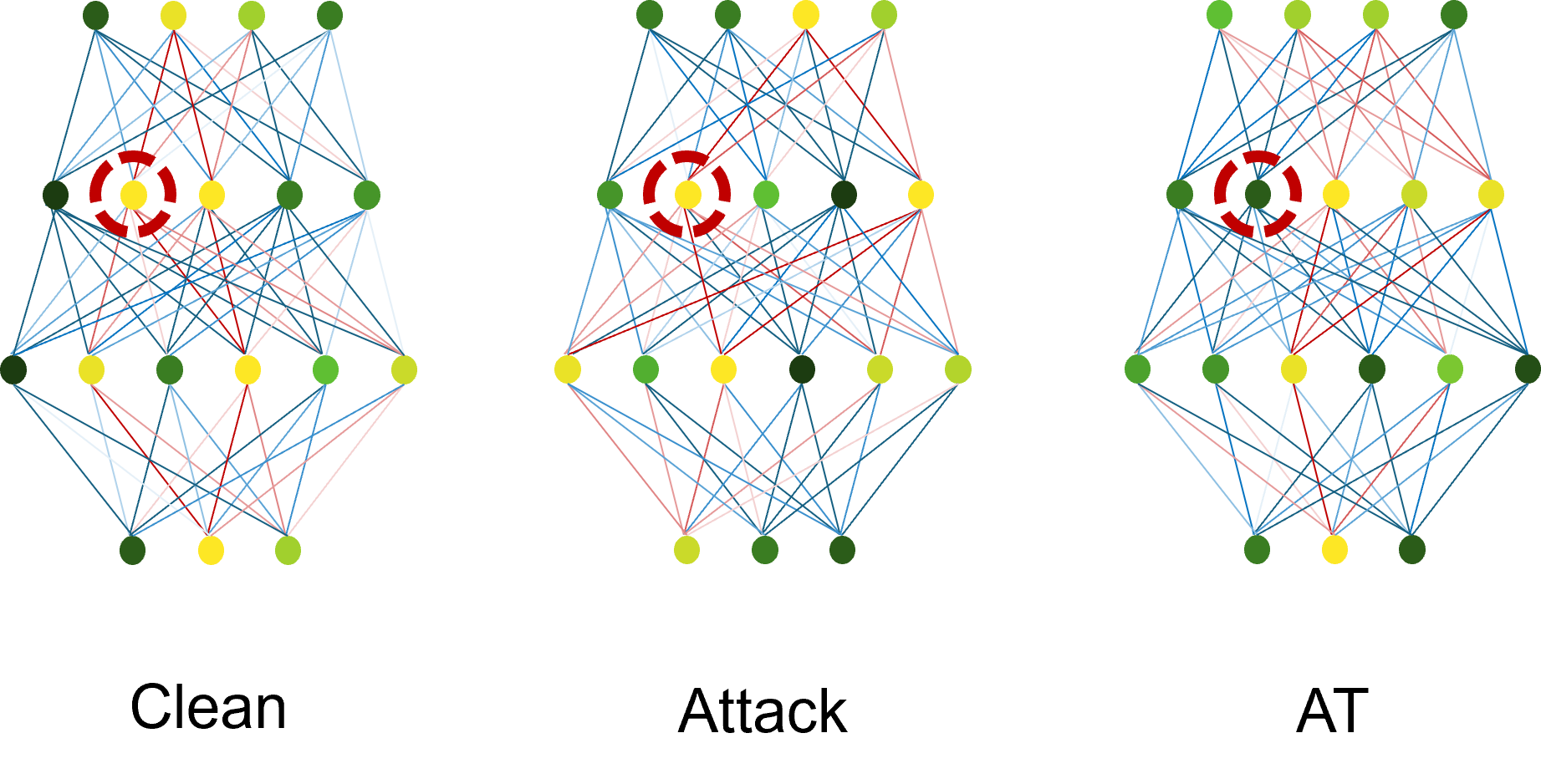}
    \caption{Importance score of models with different inputs.}
    \label{fig:importance-a}
  \end{subfigure}
  \hspace{0.01\linewidth}
  \vrule width 0.1pt
  \hspace{0.01\linewidth}
  \begin{subfigure}{0.60\linewidth}
  \includegraphics[width=1\linewidth]{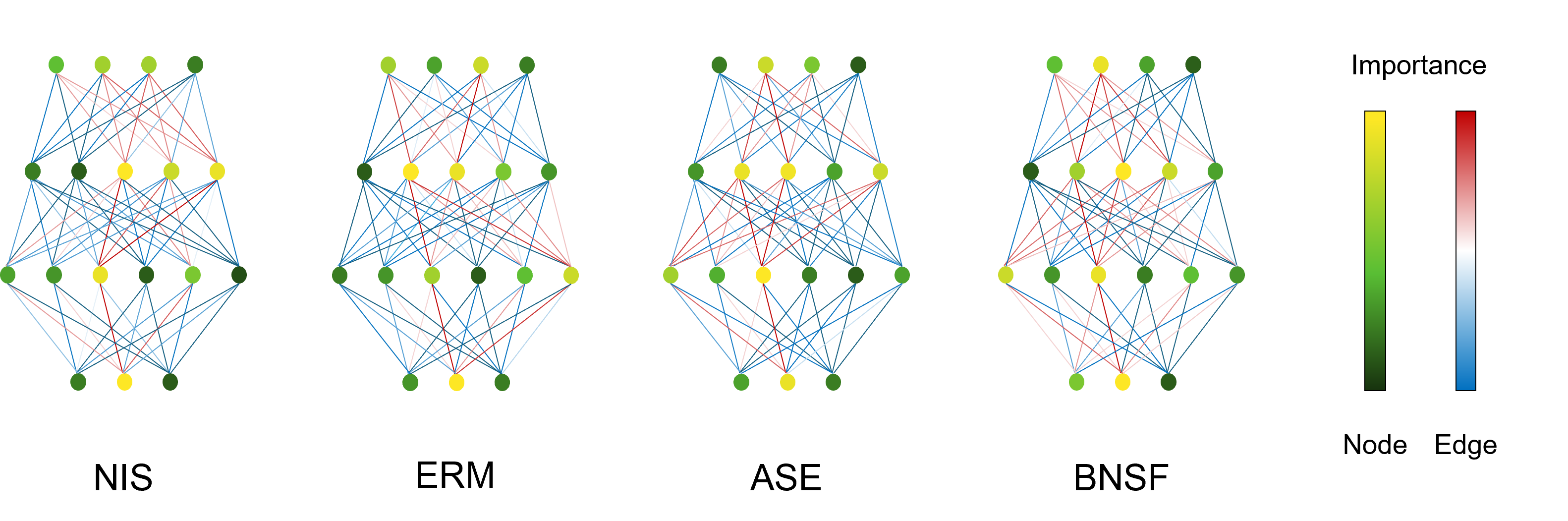}
    \caption{Importance score of a model with different scoring metrics.}
    \label{fig:importance-b}
  \end{subfigure}
  \caption{Importance score visualization for the 4-5-6-3 CNN model on the MNIST dataset: yellow/red indicate higher importance, green/blue indicate lower.}
  \label{fig:importance}
  \vspace{-2mm}
\end{figure*}

\noindent
\textbf{Neuron Importance Score (NIS)}
NIS~\cite{yu2018nisp} is a common pruning score that minimizes the weighted distance between the original and pruned final responses in a given layer.
Given a neural network \( f^{(n)} \) with \( n \) layers, and the importance score of the final layer response \( I^{(n)} \), the importance score of the \( i \)-th layer \( I^{(i)} \) is calculated as follows:
\begin{equation}
    I^{(i)} = {|w^{(i+1)}|}^T \cdot {|w^{(i+2)}|}^T \cdots {|w^{(n)}|}^T \cdot I^{(n)}
\end{equation}
Here, \( w^{(i)} \) refers to the weight matrix of the \( i \)-th layer. In this case, the importance of a neuron in the network can be computed recursively along the network. Thus, \( I^{(i)} \) can be propagated from the importance score of the \( (i+1) \)-th layer as follow:
\begin{equation}
    I_{NIS}^{(i)} = {|w^{(i+1)}|}^T \cdot I^{(i+1)}
\end{equation}

\noindent
\textbf{Empirical Risk Minimization (ERM)} 
The importance score based on ERM, used in HYDRA\cite{sehwag2020hydra} and HARP, is defined as follows, based on the weight magnitudes of a pre-trained model.
\begin{equation}
    I^{(i)}_{ERM} = \frac{\eta \cdot |\theta^{(i)}|}{\max(|\theta^{(i)}|)}
\end{equation}
Here,\( I^{(i)}_{ERM} \) represents the importance score for the weights of the \( i \)-th layer, \( \theta^{(i)} \) denotes the weight matrix of that layer, and \( \eta \) is a scaling factor determined by the layer's structure and receptive field size.

\noindent
\textbf{Adversarial Saliency Estimation (ASE)} 
MAD~\cite{lee2022masking} computes the ASE for each parameter, removing the least salient ones. Adversarial saliency is estimated via a second-order Taylor expansion, with the saliency score \( I_{ASE} \) for each parameter \( w^{(i)} \) defined as:

\begin{equation}
    I_{ASE} = \frac{1}{2} \left[ \frac{\partial^2 \mathcal{L}}{\partial (w^{(i)})^2} \right] |w^{(i)}|^2
\end{equation}
Here, \( \frac{\partial^2 \mathcal{L}}{\partial (w^{(i)})^2} \) is the Hessian of the loss function with respect to \( w^{(i)} \), reflecting the impact of parameter changes on the loss, and \( |w^{(i)}|^2 \) reflects the magnitude and importance of the corresponding parameter.

\noindent
\textbf{Batch Normalization Scaling Factor (BNSF)} 
BNAP~\cite{wei2021batch} identifies the magnitude of the Batch Normalization (BN) scaling factor \( \gamma \) as a robustness measure across channels, used as the pruning criterion. The BN layer normalizes weights \( w_{ik}^j \) into the following effective weights:

\begin{equation}
    \hat{w}_{ik}^j = \frac{\gamma_i^j w_{ik}^j}{(\sigma_i^j)^2 + \epsilon} \approx \frac{\gamma_i^j w_{ik}^j}{\sigma_i^j}
\end{equation}
where $i$ is the layer index, $j$ is the channel index of $i$-th layer, and $k$ is the channel index in the $(i-1)$-th layer. In this case, \( \epsilon \) is small enough to be neglected, and \( \gamma_i^j \) and \( \sigma_i^j \) each play a role in readjusting the weights per channel. Therefore, the computed weights can be used as the importance score.


\subsection{A Closer Look at Adversarial Pruning via Importance Score}\label{sec:3.3}
Least Importance Score~\cite{yu2018nisp} is a common pruning criterion that assesses each parameter's (weights, filters, channels) importance to overall network performance, pruning the least important first. This score clarifies the impact of adversarial examples on pruning. We compared importance scores between models trained on clean data, adversarial data, and adversarially-trained models using a simple CNN architecture.

As shown in Fig.~\ref{fig:importance-a}, importance scores differ significantly between models trained on clean versus adversarial datasets. In the AT model, trained on both, scores do not consistently fall between clean and adversarial values but more closely resemble the clean model. Notably, nodes highly important in both cases (e.g., the second node in the second layer) may appear less important in the AT model. This suggests that training on both datasets may obscure importance, potentially leading to pruning of critical nodes or connections.

Moreover, we evaluated the same model with different importance scoring methods mentioned earlier. The results in Fig.\ref{fig:importance-b} suggest that the order of importance for nodes and connections can vary depending on the scoring method. In other words, each sub-network pruned with a different importance scoring method may contain important information that is not shared by the others. This implies that sub-models sharing the same base model could act as multiple models in ED, while complementing each other.

\section{Efficient Ensemble Defense (EED)}

A common hypothesis in adversarial ensemble defense is that training and combining multiple base models enhances adversarial robustness compared to a single model \cite{deng2024understanding}. However, traditional ED methods often increase model capacity, which is counterproductive for reducing computational and memory costs. But is it truly impossible to achieve an efficient adversarial ensemble defense for a compact model?

We propose a novel Efficient Ensemble Defense (EED) method that leverages model compression while maintaining robustness. Sec.\ref{sec:3.3} has demonstrated the potential for sub-models, pruned based on different importance scores from a single base model, to act as multiple base models in an ensemble. By ensembling these sub-models, we can reduce model capacity compared to traditional ED. However, it is observed that the smaller the model during ensemble, the more the robustness performance decreases; thus, high diversity is required to mitigate this issue\cite{pang2019improving}. Therefore, we enhance the ensemble effect and maintain robustness through robust diversity evaluation and misclassification regularization. Furthermore, to improve inference speed—a primary goal of model compression—we dynamically add sub-models during inference and perform ensemble only as needed, thus enhancing inference efficiency.

EED is designed with three main considerations: 1) the diversity of pruned sub-models, 2) improvement of robustness during the ensemble process, and 3) maintaining compactness throughout the ensemble process.

\begin{algorithm}[t]
    \caption{Efficient Ensemble Defense (EED)}
    \label{alg:eed}
    \begin{algorithmic}
        \STATE \textbf{Input:} Training data \(D\), base model \(M\)
        \STATE \textbf{Output:} Ensemble \(Selected\)


        \FOR{each subset \(D_i \subset D\)}
            \STATE \(F_i \gets \text{AdversarialPruning}(M, D_i)\)
            \STATE \(Set(D) \gets Set(D) \cup F_i\)
        \ENDFOR

        \STATE \(EnsSet \gets \text{FormEnsembleSet}(Set(D))\)

        \FOR{each ensemble $E \in EnsSet$}
            \STATE \(RD \gets \text{CalculateRobustDiversity}(E)\)
        \ENDFOR

        \STATE \(Selected \gets \text{SelectEnsemble}(EnsSet, RD)\)

        \FOR{each training example  $(x, y) \in D$}
            \STATE \(\mathcal{L}_{EED}(x, y) \gets \mathcal{L}_E + \mathcal{L}_R + \mathcal{L}_C\)
            \STATE \(Selected \gets \text{UpdateModels}(Selected, \mathcal{L}_{EED})\)
        \ENDFOR
        
        \COMMENT{------------ \textbf{INFERENCE PHASE} ------------}
        \FOR{each test example \(x \in T\)}
            \STATE \(pred \gets \text{InitializePredictions}(Selected)\)
            \STATE \(t \gets 1\)

            \WHILE{not sufficient robustness}
                \STATE \(pred \gets \text{UpdateEnsemble}(pred, Selected[t])\)
                \STATE \(t \gets t + 1\)
                \STATE \(q_t \gets \text{CalculateStoppingCriterion}(pred)\)
                \IF{robustness criterion met}
                    \STATE \text{break}
                \ENDIF
            \ENDWHILE
        \ENDFOR
        \STATE \textbf{Output } \(pred\)
    \end{algorithmic}
\end{algorithm}
\subsection{Enhancing Robust Diversity} \label{sec:4.1}

Our strategy begins with generating sub-models from a base model through pruning, where each sub-model undergoes an AP process based on a unique importance score. This AP adheres to the principles described in Sec.\ref{sec:3.1}, utilizing four importance scoring methods detailed in Sec.\ref{sec:3.2}. To further enhance diversity, we divide the training dataset for pruning into multiple subsets, with each subset designated to train a specific sub-model. Some subsets are shared across all sub-models to retain common core knowledge, while the remaining subsets are distributed across different sub-models to ensure each to learn somewhat different aspects of the data. These subsets are randomly reshuffled whenever the importance scoring method changes. This approach amplifies model diversity, as each sub-model becomes specialized in defending against attacks targeting specific patterns or features in the data. Ultimately, we establish a model pool composed of sub-models with varying importance scoring methods or trained on different datasets.

Let this be denoted as pool of \( N \) models trained on the dataset \( D \), represented as \( Set(D) = \{F_0, \ldots, F_{N-1}\} \). The ensemble set \( EnsSet \) represents the collection of all possible ensemble teams with sizes \( S \) ranging from 2 to \( N \) formed from \( Set(D) \). The number of ensemble teams is given by $|EnsSet| = \sum_{S=2}^{N} \binom{N}{S} = 2^N - (1 + N)$.

To quantify the diversity among sub-models within each ensemble team, we define a Robust Diversity (RD) score based on the probability of defense failure for the model. The diversity score relies on two probabilities:\( p_{one} \) and \( p_{two} \). \( p_{one} \) is the probability that a randomly chosen model fails to defend (viz. the attack succeeds), while \( p_{two} \) is the expected probability that two randomly selected models both fail to defend:
\begin{equation}
    p_{one} = \sum_{i=1}^{S} \frac{S_i}{S} p_i, \quad p_{two} = \sum_{i=1}^{S} \frac{S_i (i - 1)}{S(S - 1)} p_i
\end{equation}
Here, \( p_i \) represents the defense failure probability of the \( i \)-th model among the total \( S \) models. Based on the two probabilities, the RD score is defined by:
\begin{equation}
    RD = 1 - \frac{p_{two}}{p_{one}}
\end{equation}
This RD metric measures diversity. \( RD = 1 \) indicates maximum diversity (i.e., even if one model fails in defense, the likelihood of other models succeeding is high), while \( RD = 0 \) indicates no diversity (i.e., both models fail in defense). The EED framework applies RD-based ensemble selection to form the smallest ensemble from the \( EnsSet \) of sub-models with the high RD scores.

\subsection{Regularization against Misclassification}\label{sec:4.2}

As in the existing ED technique \cite{yang2020dverge}, the classifier is trained by combining the classification loss, regularization(\(Reg\)) and diversity(\(Div\)) terms. Specifically, for a single training example \((x, y)\), the ensemble loss function is defined by:
\begin{equation}
    \begin{split}
            \mathcal{L}_E(x, y) = \sum_{i=1}^{N} \ell(F_i(x), y) + \alpha  Reg(h(x)) \\+ \beta Div(h^1(x), h^2(x), \ldots, h^N(x), y)
    \end{split}
\end{equation}
Here, \( \alpha \) and \( \beta \) are hyperparameters, and \( h^i(x) = (h_1^i(x), \dots, h_C^i(x)), i \in \mathbb{N} \) represents the output of the \( i \)-th classifier. The final output is calculated using the average combiner \( h(x) = \frac{1}{N} \sum_{i=1}^{N} h^i(x) \).

To improve the robustness of EED, we employ an additional regularization term that focuses on the probability scores of the most misclassified class, inspired by iGAT \cite{deng2024understanding}. For a given example \((x, y)\), the proposed term is defined by:
\begin{equation}
    \begin{split}
        \mathcal{L}_R(x,y) = -\psi_{0/1}(c(h^1(x), \ldots, h^N(x)), y) \\ \cdot \log \left(1 - C \max_{i=1}^{N} \max_{j=1}^{C} h_i^j(x)\right)
    \end{split}
\end{equation}
The error function \( \psi_{0/1}(f, y) = \frac{1}{|D|} \sum_{x \in D} 1\left[f_{y}(x) < \max_{c=y} f_c(x)\right] \) returns 0 if \( f \) correctly predicts the label \( y \), and 1 otherwise. This regularization term plays a crucial role in guiding the model to reduce misclassifications, especially improving its ability to handle ambiguous data points or difficult input examples. By doing so, each base classifier focuses more on the examples it misclassified, enhancing the robustness of the overall ensemble model across diverse input spaces.

\subsection{Regularization for Compactness}\label{sec:4.3}

Another key objective of the EED framework is maintaining the compressed models' compactness while maximizing the ensemble merit. To achieve this, we design a compactness regularization term \( \mathcal{L}_C \) to maintain weight and activation sparsity among the sub-models, thereby improving the compactness of the ensemble. The weight regularization term controls the sparsity of the sub-model weights, ensuring that they remain sparse above a certain threshold, thus suppressing model capacity growth. Specifically, the weight sparsity regularization term \( \mathcal{L}_W \) is defined as follows:
\begin{equation}
    \mathcal{L}_{W}(x,y) = \sum_{i=1}^{N} \| W_i \|_1,
\end{equation}
Here, \( \| W_i \|_1 \) denotes the \( L_1 \)-norm of the weights \( W_i \) of the \( i \)-th sub-model, which encourages sparsity by minimizing the sum of the absolute values of the weights.

To control the activation sparsity of the sub-models, we use an additional sparsity regularization term \( \mathcal{L}_{A} \), which encourages each sub-model to reduce unnecessary neuron activations. This term is defined as follows:
\begin{equation}
    \mathcal{L}_{A}(x,y) = \sum_{i=1}^{N} \| h^i(x) \|_0,
\end{equation}
Here, \( \| h^i(x) \|_0 \) represents the sum of the number of non-zero activations in the output \( h^i(x) \) of the \( i \)-th sub-model.

Thus, \( \mathcal{L}_C \) can be summarized as follows:
\begin{equation}
    \mathcal{L}_C(x,y) = \lambda_1 \mathcal{L}_{W} + \lambda_2 \mathcal{L}_{A}
\end{equation}
The hyperparameter \( \lambda_1 \) regulates weight sparsity, with higher values leading to more sparse weights and reduced model capacity. Similarly, \( \lambda_2 \) controls activation sparsity, making neuron activations sparser, which results in a more compact model structure that saves memory and computational resources.

Finally, the overall loss function of EED is as follows:
\begin{equation}
    \mathcal{L}_{EED}(x, y) = \mathcal{L}_E(x, y) + \omega \mathcal{L}_R(x, y) + \gamma \mathcal{L}_C(x, y)
\end{equation}
The EED framework provides high robustness against adversarial attacks while ensuring compactness through these multiple regularization terms, making it suitable to efficient applications like real-time inference.

\begin{table*}[t!]
    \begin{center}
    \caption{Comparison of various AP methods and EED on CIFAR-10 and SVHN datasets when sparsity rate $s_r = 80\%$}
    \fontsize{8.0pt}{9.0pt}\selectfont
    {\setlength\tabcolsep{5.5pt} 
    \begin{tabular}{c|c|cccccc|cccccc|c}
        \toprule
        \multirow{2}{*}{ } & \multirow{2}{*}{Setting} & \multicolumn{6}{c|}{CIFAR-10} &                  \multicolumn{6}{c|}{SVHN}\\
                                            & & Clean &  PGD  &  AA   &  C\&W   &  DF   & Speed up  &  Clean &  PGD &  AA   &   C\&W   &   DF   & Speed up & Prams\\
        \midrule
        \multirow{9}{*}{Resnet 18}                                                     
        & AT\cite{madry2018PGD}               & 87.05 & 56.14 & 48.02 & 57.60 & 53.10 &   1.00x   & 93.37 & 56.27 &  50.14 & 58.85 & 59.80 &  1.00x   & 11.2M\\
        \cline{2-15}
        & R-ADMM\cite{ye2019adversarial}      & 81.25 & 48.00 & 43.92 & 49.17 & 39.11 &   1.68x   & 74.81 & 49.73 & 37.40 & 52.62 & 43.40 &  1.63x   & \multirow{9}{*}{2.2M}\\
        & HYDRA\cite{sehwag2020hydra}         & 77.36 & \underline{52.92} & 43.74 & 49.64 & 45.91 &   1.77x   & 91.06 & 52.22 & \underline{47.62} & 55.13 & 46.13 &  1.74x   \\
        & RST\cite{fu2021drawing}             & 61.02 & 41.01 & 18.38 & 51.02 & 26.82 &   \textbf{1.86x}   & 82.39 & 46.29 & 36.35 & 52.65 & 39.27 &  \underline{1.83x}   \\
        & MAD\cite{lee2022masking}            & 74.18 & 50.38 & 41.27 & 54.17 & 38.44 &   1.69x   & 92.84 & 51.65 & 39.87 & \textbf{59.80} & 44.72 &  1.70x   \\
        & Flying Bird\cite{chen2022sparsity}  & 81.07 & 51.62 & 44.41 & 56.08 & 45.29 &   1.79x   & 90.21 & 52.06 & 42.01 & 57.30 & \underline{52.23} &  1.76x   \\
        & HARP\cite{zhao2023holistic}         & \underline{83.84} & 52.56 & \underline{45.36} & 56.57 & \underline{47.04} &   1.81x   & 92.60 & \underline{54.16} & 45.89 & 57.28 & 51.24 &  1.80x   \\
        & TwinRep\cite{li2023learning}        & 77.26 & 52.04 & 43.52 & \underline{56.60} & 46.18 &   \underline{1.82x}   & \underline{92.96} & 53.83 & 44.73 & 56.26 & 48.61 &  1.78x   \\
        & \textbf{EED(ours)}                  & \textbf{86.13} & \textbf{55.71} & \textbf{48.13} & \textbf{57.03} & \textbf{51.97} &   \textbf{1.86x}   & \textbf{93.15} & \textbf{55.74} & \textbf{50.18} & \underline{58.37} & \textbf{56.05} &  \textbf{1.85x}   \\
        \midrule
        \multirow{9}{*}{VGG 16}
        & AT\cite{madry2018PGD}               & 82.70 & 54.49 & 48.52 & 54.91 & 56.91 &   1.00x   & 93.06 & 57.64 & 52.28 & 55.36 & 61.49 &  1.00x   & 14.7M\\
	  \cline{2-15}
        & R-ADMM\cite{ye2019adversarial}      & 76.51 & 47.19 & 40.41 & 44.80 & 48.29 &   1.65x   & 65.40 & 48.39 & 44.64 & 48.92 & 49.85 &  1.66x   & \multirow{9}{*}{2.9M}\\
        & HYDRA\cite{sehwag2020hydra}         & 78.83 & 47.85 & 44.85 & 45.73 & 49.41 &   1.72x   & 90.82 & 53.66 & 46.30 & 52.63 & 53.42 &  1.72x   \\
        & RST\cite{fu2021drawing}             & 77.29 & 39.90 & 30.15 & 32.02 & 42.03 &   \textbf{1.83x}   & 82.39 & 41.94 & 45.52 & 43.58 & 38.81 &  \textbf{1.81x}   \\
        & MAD\cite{lee2022masking}            & 73.65 & 48.81 & 40.88 & 47.16 & 45.50 &   \underline{1.80x}   & 89.94 & 49.46 & 40.31 & 53.03 & 47.37 &  1.69x   \\
        & Flying Bird\cite{chen2022sparsity}  & 77.36 & 49.43 & \underline{45.73} & 50.38 & 48.34 &   1.68x   & 90.11 & 53.29 & \underline{49.18} & 52.81 & 50.78  &  1.73x   \\
        & HARP\cite{zhao2023holistic}         & \underline{80.92} & \underline{51.77} & 44.02 & \underline{52.79} & \underline{52.17} &   1.73x   & \textbf{93.22} & \underline{55.53} & 44.73 & \underline{52.96} & 56.16 &  \underline{1.77x}   \\
        & TwinRep\cite{li2023learning}        & 76.04 & 50.12 & 43.26 & 52.42 & 51.59 &   1.79x   & 92.60 & 54.61 & 49.06 & 51.24 & \underline{56.58} &  1.75x   \\
        & \textbf{EED(ours)}                  & \textbf{81.39} & \textbf{54.26} & \textbf{47.49} & \textbf{53.27} & \textbf{53.94} &   1.79x   & \underline{93.14} & \textbf{57.19} & \textbf{51.49} & \textbf{53.12} & \textbf{60.27} &  \underline{1.77x}   \\
        \bottomrule
        \end{tabular}}\label{tab:main}
    \end{center}
    \vspace{-5mm}
\end{table*}
\subsection{Dynamic Ensemble for Inference Efficiency}\label{sec:4.4}
In the training phase, EED optimizes all sub-models to ensure diversity and robustness. To enhance the compactness and efficiency of EED during the inference phase, we adopt an inference-efficient ensemble method \cite{li2023towards}. We apply this to adversarial defense by dynamically constructing an ensemble of varying sizes for each sample, termed Dynamic Inference Ensemble (DIE). Specifically, for a given sample \( x \), EED dynamically determines the optimal point to stop while adding each sub-models progressively. As a result, by monitoring the uncertainty of each sub-model on adversarial examples and only introducing additional models when necessary, we avoid unnecessary computations while maintaining high robustness.

DIE sequentially adds sub-models and evaluates the robustness of the predictions at each stage. Starting with the first sub-model \( F_1 \), a new sub-model \( F_t \) is added at each stage \( t \), and the intermediate ensemble prediction \( \hat{y}_{ens}^t \) is updated. These intermediate predictions serve as a criterion for assessing how robust the added sub-models are. 
The intermediate ensemble prediction based on the predictions \( \hat{y}_i \) of each sub-model \( F_i \) is computed by:
\begin{equation}
    \hat{y}_{ens}^t = \frac{1}{t} \sum_{i=1}^{t} \hat{y}_i
\end{equation}
Here, each \( \hat{y}_{ens}^t \) represents the combined prediction of the sub-models for the input (both clean and adversarial examples), and it plays a role in strengthening the overall robustness of the ensemble.

At each stage \( t \), the early stopping probability \( q_t \) is assessed. If the current ensemble performance is determined to be robust enough, the addition of sub-models is halted. The value of \( q_t \) is computed based on the prediction uncertainty and confidence. The prediction uncertainty is computed by KL-divergence of $t$ and $(t-1)$-th ensemble predictions, and the prediction confidence setting follows WoC\cite{wang2021wisdom}.
The optimal stopping point \( z_x \) is defined as follows, based on a geometric distribution modeling with discrete Bernoulli trials:
\begin{equation}
    z_x = \arg\max_t \left( q_t \prod_{i=1}^{t-1} (1 - q_i) \right)
\end{equation}
Here, \(q_t\) is dynamically adjusted at each stage based on robustness, and sub-models are added only when necessary, allowing for efficient use of computational resources.

The final ensemble prediction \( \hat{y}_{ens} \) is calculated as the average of the predictions of all sub-models up to the stopping point \( z_x \):
\begin{equation}
    \hat{y}_{ens} = \frac{1}{z_x} \sum_{i=1}^{z_x} \hat{y}_i
\end{equation}
This approach takes into account the predictions of all sub-models while avoiding the inclusion of unnecessary ones, thereby enhancing efficiency. By applying different ensemble sizes for each sample, the overall computational cost is reduced while still being robust against adversarial attacks.

DIE finds a balance between efficiency and robustness, quickly identifying the optimal combination of sub-models to ensure strong performance even in real-time inference settings. As a result, the EED framework can lower the average inference cost in adversarial attack scenarios while maintaining high level of robustness, making it an effective defense, even in resource-constrained environments.

\section{Experimental Evaluations}

\subsection{Experiment Setups} \label{sec:5.1}
\noindent
\textbf{Classifiers and Datasets}
We utilize two specific architectures, ResNet18 and VGG16, as classifiers and conduct experiments focusing on CIFAR-10 and SVHN. We divide the dataset into \( d \) subsets for pruning into various sub-models, setting \( d = 4 \) to create three sub-models for each pruning metric. 25\% of the dataset is shared across all sub-models, while the remaining 75\% is allocated separately for each of the three sub-models. With four pruning metrics in play, we end up with a total of \( N = 12 \) sub-models, which together form the ensemble set, denoted as \( EnsSet \). For ensemble learning, the entire dataset is used.

\noindent
\textbf{Attacks}
We use PGD-20 attacker with a step size of 2/255 to generate adversarial examples for training, and Auto Attack(AA)\cite{croce2020reliable}, C\&W\cite{carlini2017towards} and Deepfool(DF)\cite{moosavi2016deepfool} for evaluation. Except for DF using $l_2$ norm, the rest use $l_\infty$ norm, and all attacks use perturbation strength $\epsilon$ = 8/255.

\noindent
\textbf{Training}
We use the Madry\cite{madry2018PGD} as AT method. The batch size is set to 512, and the multi-step learning rates are \{0.01, 0.002\} for CIFAR-10 and \{0.1, 0.02, 0.004\} for SVHN. The hyperparameters are set as follows: \( \omega = 10 \), \( \lambda_1 = 0.7 \), \( \lambda_2 = 0.25 \), and the RD threshold is set to 0.7. $c_r$ is set to be more than four times sparser than the desirable $s_r$, considering the size of $N$, and $\gamma$ is adjusted to ensure that the $s_r$ is met even after the ensemble. (When $s_r$=80\%, we set $c_r=95\%$ and $\gamma=4$.)
Each experiment is run on an NVIDIA RTX A6000 GPU with 8 CPU cores.

\begin{figure*}[t!]
    \centering
    \begin{minipage}{0.27\textwidth}
        \centering
        \includegraphics[width=\textwidth]{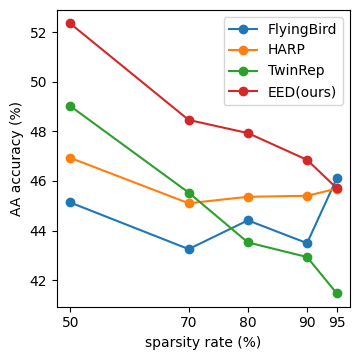}
        \caption{Robust performance by $s_r$.}
        \label{fig:sparsity}
    \end{minipage}
    \hspace{0.05\textwidth} 
    \begin{minipage}{0.66\textwidth}
        \centering
        \begin{subfigure}{0.48\linewidth}
            \includegraphics[width=1\linewidth]{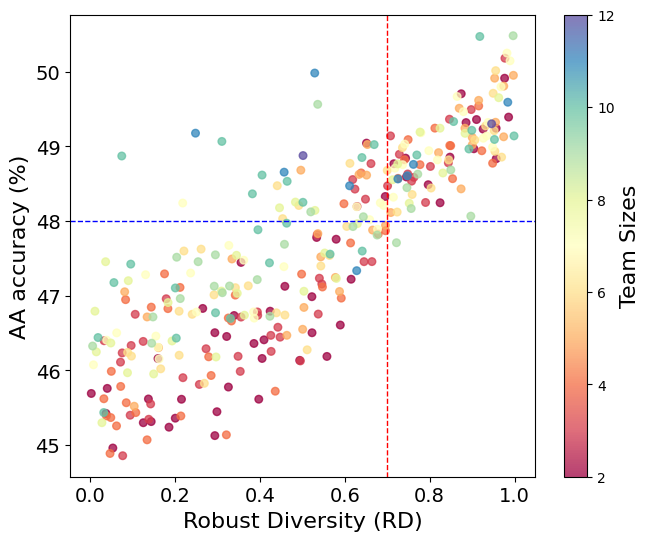}
            \caption{RD analysis on ensemble team size.}
            \label{fig:RD1}
        \end{subfigure}
        \hfill
        \begin{subfigure}{0.48\linewidth}
            \includegraphics[width=1\linewidth]{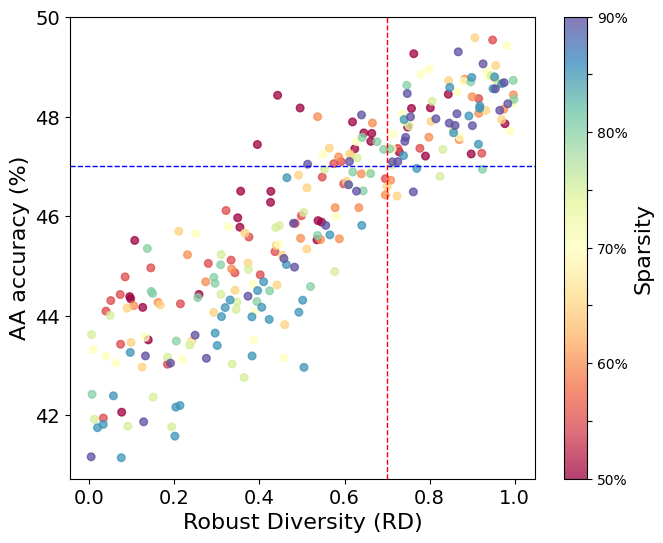}
            \caption{RD analysis on sub-model sparsity.}
            \label{fig:RD2}
        \end{subfigure}
        \vspace{-1mm}
        \caption{Correlation between RD and robustness in different ensemble components.}
        \label{fig:RD}
    \end{minipage}
    \vspace{-1mm}
\end{figure*}
\subsection{Result Comparison and Analysis} \label{sec:5.2}

This section compares the proposed EED with existing APs for sparsity rate(\(s_r\)) of 80\% on the CIFAR-10 and SVHN datasets. Table \ref{tab:main} demonstrates that EED consistently provided superior adversarial robustness, with significantly higher accuracy against attacks such as PGD, AA, C\&W, and DF compared to other methods. Additionally, EED maintained high performance on clean data, indicating a proper balance between adversarial robustness and clean accuracy.

Note that EED demonstrates superior robustness, particularly against strong attacks such as AA and PGD, outperforming existing defense methods like HYDRA, HARP, and TwinRep. For example, in the CIFAR-10 experiment using the ResNet-18 architecture, EED achieved 55.71\% accuracy against PGD and 48.13\% against AA, showing an improvement of over 2\% compared to other methods and achieving the highest robustness. Moreover, despite using only 20\% of the model's parameters, the performance drop in AT was minimal, less than 0.5\%. Also, EED showed stronger defense capabilities than AT against PGD in the ResNet-18-based SVHN experiment. 

As a result, EED accomplished remarkable performance improvements without sacrificing efficiency. Notably, it also records up to 1.86 times faster inference speeds, comparable to or exceeding existing techniques. This is crucial for practical applications where computational resources may be limited, demonstrating that EED is an ideal defense method for resource-constrained environments. Furthermore, EED showed consistent performance across different model architectures such as ResNet-18 and VGG-16, suggesting that EED can be applied to a wide range of network architectures with minimal adjustment.

\subsection{Ablation Studies} \label{sec:5.3}
\noindent
Ablation studies are conducted on Resnet18 and CIFAR-10.

\noindent
\textbf{Changes along Sparsity}
Fig.\ref{fig:sparsity} shows the change in robustness performance with respect to the sparsity rate. EED demonstrated the highest performance compared to the existing top three methods, except for a very high $s_r$ (95\%) against AA. EED tended to show a decrease in performance as $s_r$ increases, which is believed to be due to the reduction in ensemble diversity caused by the higher compression rate. On the other hand, at lower $s_r$, EED exhibited higher robustness than AT due to the enhanced ensemble diversity. More detailed experimental results for different $s_r$ can be found in the supplementary materials.

\noindent
\textbf{Analyzing Robust Diversity}
Fig.\ref{fig:RD} analyzes the role of RD in EED. Fig.\ref{fig:RD1}, which shows robustness according to ensemble team size, demonstrates that a high RD is necessary for small ensemble sizes to achieve high performance. Fig.\ref{fig:RD2} shows robustness from the RD perspective with respect to the compression rate ($c_r$) of the sub-models. We can find that to maintain high performance while keeping sub-models small, a high RD is also required.

\noindent
\textbf{Analyzing Regularization Components}
EED has two regularization terms, as mentioned in Sec.\ref{sec:4.2} and Sec.\ref{sec:4.3}. Regularization against misclassification (${L}_R$) is controlled by $\omega$, while regularization for compactness (${L}_C$) is controlled by $\gamma$. This section analyzes the impact of each term on the performance of EED. Tab.\ref{tab:reg} shows that ${L}_R$ has a greater effect on clean performance, while ${L}_C$ has a larger impact on attack performance. In particular, when the compactness term is not used, attack performance significantly deteriorates, which is likely due to the need of reducing the ensemble size further to meet the $s_r$, thereby diminishing the ensemble advantages.

\noindent
\textbf{Effect of Dynamic Inference Ensemble}
Tab.\ref{tab:dynamic} analyzes the effect of DIE for different sparsity rates. At $s_r=80\%$, DIE had little impact on the main performance and only showed a slight speed improvement. On the other hand, at $s_r=50\%$, activating DIE resulted in performance improvements of 1.81\% and 1.59\% for PGD and AA, respectively, while significantly improving inference speed. This suggests that DIE is more effective at lower sparsity. The reason for this performance improvement is that, with lower sparsity, the ensemble can be larger, thus enhancing the optimization effect of the dynamic ensemble. In contrast, without DIE, the likelihood of conflicts between the sub-models may increase as the ensemble grows larger.

\begin{table}[t!]
    \begin{center}
    \caption{Analysis of EED with regularization terms.}
    \vspace{-1mm}
    \fontsize{8.0pt}{9.0pt}\selectfont
    {\setlength\tabcolsep{5.5pt} 
    \begin{tabular}{c|c|ccc}
        \toprule
                             & Settings & Clean &  PGD  &  AA\\
        \midrule
        \multirow{2}{*}{CIFAR-10}                                                     
        & EED                           & 86.13 & 55.71 & 48.13 \\
        \cline{2-5}
        & $\omega=0$                    & 74.16 & 47.45 & 36.37 \\
        \multirow{2}{*}{Resnet 18}
        & $\gamma=0$                    & 76.33 & 39.10 & 31.59 \\
        & $\omega,\gamma=0$             & 73.95 & 37.83 & 28.94 \\
        \bottomrule
        \end{tabular}}\label{tab:reg}
    \end{center}
    \vspace{-4mm}
\end{table}

\begin{table}[t!]
    \begin{center}
    \caption{Correlation between dynamic inference ensemble and $s_r$.}
    \vspace{-1mm}
    \fontsize{8.0pt}{9.0pt}\selectfont
    {\setlength\tabcolsep{5.5pt} 
    \begin{tabular}{c|c|cccc}
        \toprule
                             & Settings & Clean &  PGD  &  AA    & Speed up\\
        \midrule
        \multirow{2}{*}{$s_r=80\%$}                                                     
        & w/ dynamic                     & 86.13 & 55.71 & 48.13 & 1.86x \\
        & w/o dynamic                    & 86.20 & 55.62 & 48.11 & 1.80x\\
        \midrule
        \multirow{2}{*}{$s_r=50\%$}
        & w/ dynamic                     & 86.15 & 56.09 & 52.35 & 1.64x\\
        & w/o dynamic                    & 85.93 & 54.28 & 50.76 & 1.27x\\
        \bottomrule
        \end{tabular}}\label{tab:dynamic}
    \end{center}
    \vspace{-5mm}
\end{table}
\section{Conclusion}
This paper proposes the Efficient Ensemble Defense (EED) technique to maintain both compression and robustness of deep learning models. EED constructs an ensemble of sub-models using different pruning importance scores to enhance adversarial robustness while reducing model size and computational cost. By addressing information loss and limited model diversity caused by pruning with ensemble methods and regularization, EED outperforms existing AP methods in accuracy and robustness. Experimental results show superior resilience of EED against various attacks on the CIFAR-10 and SVHN, maintaining inference efficiency. EED also proves its applicability in diverse neural network architectures and resource-constrained environments.
\section*{Acknowledgment}
This work was supported by Institute of Information \& communications Technology Planning \& Evaluation (IITP) grant funded by the Korea government (MSIT) (No.RS-2022-00155915 (Artificial Intelligence Convergence Innovation Human Resources Development (Inha University)) and No. 2021-0-02068 (AI Innovation Hub)), and this work was partly supported by the National Research Foundation of Korea (NRF) grant funded by the MSIT (No. 2022R1A2C2010095).
{
    \small
    \bibliographystyle{ieeenat_fullname}

}

\clearpage
\setcounter{page}{1}
\setcounter{section}{0}
\maketitlesupplementary

\renewcommand{\thesection}{\Alph{section}}

\section{Implementation Details}\label{sec:apdx_implementation}

\subsection{Additional Experimental Setups}
We used VGG16 and ResNet18 architectures for experiments on small-scale datasets, CIFAR-10 and SVHN. Also, we employed ResNet50 as the classifier for additional large-scale datasets, ImageNet and CIFAR-100. The hyperparameter $\varphi$ was set to 0.01 for small-scale datasets and 0.1 for large-scale datasets.  
As adversarial training methods, we adopted Madry-AT~\cite{madry2018PGD}, TRADES-AT~\cite{zhang2019theoretically}, and MART-AT~\cite{wang2019improving}. For TRADES-AT and MART-AT, the original regularization values of $\lambda = 6.0$ and $\lambda = 5.0$, respectively, were set.  
During training, we allocated 100 epochs for AT (pre-training), 20 epochs for pruning, 40 epochs for sub-model fine-tuning, and 80 epochs for ensemble learning.

\subsection{Detailed Descriptions on Adversarial Pruning}
As described in \cref{sec:3.1}, following HARP~\cite{zhao2023holistic}, we dynamically learned the optimal layer-wise compression rates along with importance scores to construct a pruning mask \( M \) that maximizes adversarial robustness while satisfying the target sparsity constraint. The core objective is formulated as a min-max optimization problem, as shown in \cref{eq:pruning}.  
Here, \(\mathcal{L}_r\) represents the adversarial robustness loss based on adversarial training (AT), and \(\mathcal{L}_p\) is the global compression control loss that ensures the overall sparsity aligns with the target compression rate.

To explain the pruning process in greater detail, each layer \( l \) is associated with a binary pruning mask \( M^{(l)} \), which determines the pruned parameters. This mask is computed using a learnable importance score matrix \( I^{(l)} \) and the layer-wise compression rate \( c^{(l)} \). A pruning threshold \( a^{(l)} = 1 - c^{(l)} \) is derived, and parameters with importance scores below this threshold are pruned:
\begin{equation}
   M^{(l)} := \mathbf{1}\left[I^{(l)} > P(a^{(l)}, I^{(l)})\right] 
\end{equation}
Here, \( P(a^{(l)}, I^{(l)}) \) represents the percentile calculated based on the pruning threshold \( a^{(l)} \) and the importance scores \( I^{(l)} \). The compression rate of each layer \( c^{(l)} \) is parameterized using a learnable compression quota \( r^{(l)} \) and a sigmoid-based activation function \( g(r) \). The resulting values are constrained within the range \([a_{\text{min}}, 1]\), as expressed below:
\begin{equation}
    a^{(l)} = g(r^{(l)}) = (1 - a_{\text{min}}) \cdot \text{sig}(r^{(l)}) + a_{\text{min}}
\end{equation}
Such dynamic control prevents excessive pruning of any specific layer by ensuring that the pruning threshold satisfies \( a^{(l)} \geq a_{\text{min}} \).

To align the overall compression ratio with the target sparsity \( a_t \), the compression loss \( \mathcal{L}_p \) adjusts the deviation as follows:  
\begin{equation}
    L_p(\theta \odot M, a_t) := \max\left(\frac{\|\theta_{\neq 0}\|_1}{a_t \cdot \|\theta\|_1} - 1, 0\right)
\end{equation}
where \( \|\theta_{\neq 0}\|_1 \) represents the number of non-zero parameters remaining after pruning.

\subsection{Detailed Descriptions on Dynamic Inference Ensemble}
As mentioned in \cref{sec:4.4}, the core of Dynamic Inference Ensemble (DIE) is to determine the optimal stopping point \( z_x \) for a given sample \( x \), maximizing robustness while ensuring computational efficiency.  
In more detail, the probability that all models remain active up to and including step \( t \) is defined as:
\begin{equation}
    S(t) = 1 - \sum_{i=1}^{t-1} q_i
\end{equation}
which represents the likelihood that all models have participated in inference until step \( t \). The conditional probability \( q_t \) of stopping inference at step \( t \), given that inference has not stopped prior, is expressed as:
\begin{equation}
    q_t = \frac{S(t) - S(t + 1)}{S(t)}
\end{equation}
The sequential decision-making process at each step aims to optimize the probability \( z_t \) to identify \( z_x \), where \( z_t \) is derived as:
\begin{equation}
    z_t = q_t \prod_{i=1}^{t-1} (1 - q_i)
\end{equation}

\begin{table*}[t]
    \begin{center}
    \caption{Comparison of various AP methods and EED on CIFAR-100 and ImageNet datasets when sparsity rate $s_r = 90\%$}
    \fontsize{8.0pt}{9.0pt}\selectfont
    {\setlength\tabcolsep{5.5pt} 
    \begin{tabular}{c|c|cccccc|cccccc|c}
        \toprule
        \multirow{2}{*}{ } & \multirow{2}{*}{Setting} & \multicolumn{6}{c|}{CIFAR-100} &                  \multicolumn{6}{c|}{Imagenet}\\
                                            & & Clean &  PGD  &  AA   &  C\&W   &  DF   & Speed up & Clean &  PGD &  AA   &  C\&W  &  DF   & Speed up & Prams\\
        \midrule
        \multirow{9}{*}{Resnet 50}                                                     
        & AT\cite{madry2018PGD}               & 64.37 & 36.29 & 28.07 & 30.18 & 39.80 &   1.00x    & 60.25 & 32.38 & 28.79 & 30.67 & 34.54 &  1.00x   & 25.6M\\
        \cline{2-15}
        & R-ADMM\cite{ye2019adversarial}      & 61.38 & 31.23 & 21.85 & 24.04 & 30.42 &   2.43x    & 35.26 & 14.35 & 11.01 & 12.35 & 13.53 &  2.41x   & \multirow{9}{*}{2.6M}\\
        & HYDRA\cite{sehwag2020hydra}         & 62.10 & 33.52 & 24.12 & 26.20 & 33.94 &   2.80x    & 49.44 & 23.75 & 19.88 & 21.60 & 23.14 &  2.73x   \\
        & RST\cite{fu2021drawing}             & 61.14 & 29.81 & 20.15 & 21.38 & 28.45 &   \textbf{2.98x}    & 27.09 & 12.23 & 10.09 & 11.22 & 12.34 &  \textbf{2.96x}   \\
        & MAD\cite{lee2022masking}            & 56.88 & 30.59 & 21.53 & 23.55 & 30.97 &   2.55x    & 34.62 & 14.67 & 11.24 & 12.42 & 13.47 &  2.60x   \\
        & Flying Bird\cite{chen2022sparsity}  & 60.03 & 33.18 & 24.91 & 24.19 & 32.26 &   2.75x    & 48.39 & 23.05 & 18.14 & 19.33 & 17.65 &  2.56x   \\
        & HARP\cite{zhao2023holistic}         & \underline{62.51} & 33.40 & \underline{25.36} & 27.25 & \underline{34.20} &   2.87x    & \underline{55.21} & \underline{27.10} & 22.57 & \underline{24.62} & \underline{25.57} &  \underline{2.88x}   \\
        & TwinRep\cite{li2023learning}        & 62.31 & \underline{34.08} & 24.44 & \textbf{28.92} & 33.73 &   2.86x    & 52.46 & 26.73 & \underline{24.75} & 24.37 & 24.68 &  2.67x   \\
        & \textbf{EED(ours)}                  & \textbf{63.60} & \textbf{36.29} & \textbf{26.79} & \underline{28.01} & \textbf{37.14} &   \underline{2.90x}   & \textbf{58.41} & \textbf{30.54} & \textbf{26.89} & \textbf{27.15} & \textbf{29.92} &  2.84x   \\

        \midrule
        \multirow{5}{*}{Resnet 101}
        & AT\cite{madry2018PGD}               & - & - & - & - & - & - & 83.29 & 62.23 & 51.34 & 58.65 & 47.29  &  1.00x   & 44.6M\\
        \cline{2-15}
        & Flying Bird\cite{chen2022sparsity}  & - & - & - & - & - & - & 72.56 & 61.32 & 48.94 & 57.73 & \underline{48.24} &  2.61x   & \multirow{4}{*}{4.5M}\\
        & HARP\cite{zhao2023holistic}         & - & - & - & - & - & - & \underline{73.44} & \underline{62.78} & \underline{50.73} & 58.86 & 48.16 &  2.63x   \\
        & TwinRep\cite{li2023learning}        & - & - & - & - & - & - & 72.72 & 62.14 & 49.46 & 56.47 & 47.77 &  \textbf{2.80x}   \\
        & \textbf{EED(ours)}                  & - & - & - & - & - & - & \textbf{74.10} & \textbf{64.66} & \textbf{52.42} & \textbf{60.92} & \textbf{51.83} &  \underline{2.79x}   \\

        \bottomrule
        \end{tabular}}\label{tab:large}
    \end{center}
    \vspace{-5mm}
\end{table*}

\begin{table*}[t!]
    \begin{center}
    \caption{Comparison of various AP methods and EED via sparsity rate on ResNet18 architecture.}
    \fontsize{8.0pt}{9.0pt}\selectfont
    {\setlength\tabcolsep{5.5pt} 
    \begin{tabular}{c|c|cccccc|cccccc|c}
        \toprule
        \multirow{2}{*}{ } & \multirow{2}{*}{Setting} & \multicolumn{6}{c|}{CIFAR-10} &                  \multicolumn{6}{c|}{SVHN}\\
                                            & & Clean &  PGD  &  AA   &  C\&W   &  DF   & Speed up  &  Clean &  PGD &  AA   &   C\&W   &   DF   & Speed up & Prams\\
        \midrule                                                     
        & AT\cite{madry2018PGD}               & 87.05 & 56.14 & 48.02 & 57.60 & 53.10 &   1.00x   & 93.37 & 56.27 &  50.14 & 58.85 & 59.80 &  1.00x   & 11.2M\\
        \midrule
        \multirow{8}{*}{$s_r = 50\%$}
        & R-ADMM\cite{ye2019adversarial}      & 82.14 & 54.10 & 46.23 & 56.27 & 45.80 &   1.36x   & 82.40 & 49.53 & 44.52 & 50.12 & 41.05 &  1.34x   & \multirow{9}{*}{5.6M}\\
        & HYDRA\cite{sehwag2020hydra}         & 85.67 & 54.58 & 47.12 & 56.98 & 50.30 &   1.28x   & 91.12 & 53.47 & 45.83 & 50.27 & 46.58 &  1.26x   \\
        & RST\cite{fu2021drawing}             & 72.55 & 49.19 & 42.03 & 42.70 & 38.09 &   \underline{1.52x}   & 78.12 & 42.09 & 34.62 & 52.07 & 38.14 &  \underline{1.54x}   \\
        & MAD\cite{lee2022masking}            & 77.33 & 53.21 & 44.45 & 57.21 & 43.12 &   1.37x   & 92.56 & 51.04 & 42.18 & 55.34 & 49.22 &  1.31x   \\
        & Flying Bird\cite{chen2022sparsity}  & 83.02 & 53.18 & 45.13 & 57.33 & 48.02 &   1.24x   & 91.58 & 52.14 & 46.37 & 57.04 & 53.21 &  1.27x   \\
        & HARP\cite{zhao2023holistic}         & \underline{86.32} & \underline{55.29} & 46.93 & \underline{57.90} & \underline{50.41} &   1.34x   & 92.13 & \underline{53.29} & \underline{47.52} & \underline{57.26} & 54.08 &  1.36x   \\
        & TwinRep\cite{li2023learning}        & 81.45 & 54.13 & \underline{48.45} & 57.82 & 48.27 &   1.30x   & \underline{92.69} & 53.05 & 47.22 & 57.16 & \underline{54.43} &  1.32x   \\
        & \textbf{EED(ours)}                  & \textbf{88.07} & \textbf{57.83} & \textbf{52.35} & \textbf{57.92} & \textbf{53.12} &   \textbf{1.64x}   & \textbf{93.15} & \textbf{55.74} & \textbf{50.18} & \textbf{58.37} & \textbf{56.05} &  \textbf{1.63x}   \\
        \midrule
        \multirow{8}{*}{$s_r = 70\%$}
        & R-ADMM\cite{ye2019adversarial}      & 81.83 & 50.91 & 45.13 & 53.61 & 42.17 &   1.54x   & 78.24 & 50.32 & 45.21 & 51.36 & 42.28 &  1.55x   & \multirow{9}{*}{3.4M}\\
        & HYDRA\cite{sehwag2020hydra}         & 84.21 & 53.40 & \underline{46.03} & 56.40 & 47.29 &   1.56x   & 91.08 & 53.18 & 45.34 & 51.68 & 47.12 &  1.52x   \\
        & RST\cite{fu2021drawing}             & 63.89 & 42.67 & 34.31 & 35.05 & 31.52 &   \textbf{1.78x}   & 77.37 & 43.21 & 32.79 & 53.14 & 30.06 &  \textbf{1.79x}   \\
        & MAD\cite{lee2022masking}            & 75.26 & 51.85 & 42.19 & 56.70 & 40.72 &   1.68x   & 91.25 & 51.37 & 43.11 & 56.42 & 45.33 &  1.63x   \\
        & Flying Bird\cite{chen2022sparsity}  & 82.74 & 52.46 & 43.25 & 57.01 & 46.18 &   1.59x   & 91.62 & 52.09 & \underline{47.25} & 57.12 & 50.27 &  1.63x   \\
        & HARP\cite{zhao2023holistic}         & \underline{85.40} & \underline{54.71} & 45.10 & \textbf{57.56} & \underline{48.3}5 &   1.65x   & \underline{92.42} & 53.19 & 46.11 & \underline{57.28} & \underline{52.09} &  \underline{1.71x}   \\
        & TwinRep\cite{li2023learning}        & 79.82 & 52.77 & 45.53 & 57.22 & 47.01 &   1.70x   & 92.36 & \underline{53.27} & 47.13 & 56.49 & 51.43 &  1.69x   \\
        & \textbf{EED(ours)}                  & \textbf{86.45} & \textbf{56.32} & \textbf{49.16} & \underline{57.51} & \textbf{52.74} &   \underline{1.77x}   & \textbf{92.85} & \textbf{55.16} & \textbf{50.32} & \textbf{58.68} & \textbf{56.24} &  \textbf{1.79x}   \\
        \midrule
        \multirow{8}{*}{$s_r = 80\%$}                                                     
        & R-ADMM\cite{ye2019adversarial}      & 81.25 & 48.00 & 43.92 & 49.17 & 39.11 &   1.68x   & 74.81 & 49.73 & 37.40 & 52.62 & 43.40 &  1.63x   & \multirow{9}{*}{2.2M}\\
        & HYDRA\cite{sehwag2020hydra}         & 77.36 & \underline{52.92} & 43.74 & 49.64 & 45.91 &   1.77x   & 91.06 & 52.22 & \underline{47.62} & 55.13 & 46.13 &  1.74x   \\
        & RST\cite{fu2021drawing}             & 61.02 & 41.01 & 18.38 & 51.02 & 26.82 &   \textbf{1.86x}   & 82.39 & 46.29 & 36.35 & 52.65 & 39.27 &  \underline{1.83x}   \\
        & MAD\cite{lee2022masking}            & 74.18 & 50.38 & 41.27 & 54.17 & 38.44 &   1.69x   & 92.84 & 51.65 & 39.87 & \textbf{59.80} & 44.72 &  1.70x   \\
        & Flying Bird\cite{chen2022sparsity}  & 81.07 & 51.62 & 44.41 & 56.08 & 45.29 &   1.79x   & 90.21 & 52.06 & 42.01 & 57.30 & \underline{52.23} &  1.76x   \\
        & HARP\cite{zhao2023holistic}         & \underline{83.84} & 52.56 & \underline{45.36} & 56.57 & \underline{47.04} &   1.81x   & 92.60 & \underline{54.16} & 45.89 & 57.28 & 51.24 &  1.80x   \\
        & TwinRep\cite{li2023learning}        & 77.26 & 52.04 & 43.52 & \underline{56.60} & 46.18 &   \underline{1.82x}   & \underline{92.96} & 53.83 & 44.73 & 56.26 & 48.61 &  1.78x   \\
        & \textbf{EED(ours)}                  & \textbf{86.13} & \textbf{55.71} & \textbf{48.13} & \textbf{57.03} & \textbf{51.97} &   \textbf{1.86x}   & \textbf{93.15} & \textbf{55.74} & \textbf{50.18} & \underline{58.37} & \textbf{56.05} &  \textbf{1.85x}   \\
        \midrule
        \multirow{8}{*}{$s_r = 90\%$}
        & R-ADMM\cite{ye2019adversarial}      & 80.54 & 47.41 & 43.68 & 45.05 & 34.78 &   2.17x   & 84.34 & 51.91 & 38.40 & 49.46 & 42.87 &  2.23x   & \multirow{9}{*}{1.1M}\\
        & HYDRA\cite{sehwag2020hydra}         & 76.74 & 47.42 & 43.34 & 44.81 & 44.68 &   2.49x   & 88.71 & 52.49 & 44.12 & 49.61 & 45.18 &  2.43x   \\
        & RST\cite{fu2021drawing}             & 60.92 & 38.24 & 14.31 & 26.98 & 25.19 &   \textbf{2.63x}   & 74.90 & 36.43 & 34.16 & 38.03 & 38.72 &  \textbf{2.68x}   \\
        & MAD\cite{lee2022masking}            & 73.67 & 49.02 & 41.10 & 46.82 & 36.75 &   2.29x   & 89.42 & 44.29 & 37.46 & 48.90 & 41.13 &  2.31x   \\
        & Flying Bird\cite{chen2022sparsity}  & 80.69 & 51.80 & \underline{46.49} & 47.12 & \underline{49.95} &   2.42x   & \underline{91.60} & \textbf{56.36} & 39.81 & \textbf{54.01} & \underline{52.41} &  2.36x   \\
        & HARP\cite{zhao2023holistic}         & \textbf{83.38} & 50.41 & 45.40 & 47.75 & 48.53 &   2.33x   & 90.70 & 54.92 & 45.39 & 53.59 & 50.16 &  2.29x   \\
        & TwinRep\cite{li2023learning}        & 76.37 & \underline{50.82} & 42.93 & \underline{52.24} & 44.72 &   2.36x   & 88.90 & 53.06 & \underline{46.71} & 50.22 & 48.34 &  2.38x   \\
        & \textbf{EED(ours)}                  & \underline{83.26} & \textbf{52.14} & \textbf{46.85} & \textbf{56.77} & \textbf{50.21} &   \underline{2.51x}   & \textbf{91.74} & \underline{55.16} & \textbf{47.32} & \underline{53.81} & \textbf{53.57} &  \underline{2.45x}   \\
        \midrule
        \multirow{8}{*}{$s_r = 95\%$}
        & R-ADMM\cite{ye2019adversarial}      & 71.42 & 42.31 & 42.56 & 39.91 & 31.41 &   2.94x   & 64.91 & 43.55 & 36.99 & 38.38 & 40.15 &  3.07x   & \multirow{9}{*}{0.6M}\\
        & HYDRA\cite{sehwag2020hydra}         & 72.21 & 45.84 & 42.45 & 43.05 & 44.03 &   3.12x   & 85.71 & 50.56 & 45.29 & 45.53 & 46.92 &  3.19x   \\
        & RST\cite{fu2021drawing}             & 48.90 & 19.93 & 16.76 & 18.66 & 15.79 &   \textbf{3.58x}   & 61.55 & 25.90 & 23.94 & 26.39 & 28.58 &  \textbf{3.47x}   \\
        & MAD\cite{lee2022masking}            & 58.90 & 41.29 & 38.96 & 41.47 & 37.08 &   2.82x   & 86.40 & 35.62 & 24.90 & 31.79 & 37.47 &  2.88x   \\
        & Flying Bird\cite{chen2022sparsity}  & \underline{75.40} & 45.63 & \textbf{46.10} & 42.68 & 44.79 &   3.10x   & \underline{91.14} & 50.61 & \textbf{47.43} & 45.01 & 46.70 &  2.95x   \\
        & HARP\cite{zhao2023holistic}         & \textbf{77.13} & \textbf{50.41} & 45.40 & \textbf{47.75} & \textbf{48.53} &   3.39x   & \textbf{92.75} & \textbf{55.21} & 45.95 & \underline{51.3} & \textbf{47.40} &  3.19x   \\
        & TwinRep\cite{li2023learning}        & 64.97 & 47.58 & 41.47 & \underline{43.92} & 45.03 &   \underline{3.46x}   & 88.59 & 50.62 & 47.16 & \textbf{52.31} & \underline{47.35} &  \underline{3.37x}   \\
        & \textbf{EED(ours)}                  & 74.36 & \underline{48.09} & \underline{45.69} & 42.54 & \underline{46.22} &   3.27x   & 90.76 & \underline{52.74} & \underline{47.19} & 47.23 & 46.99 &  3.30x   \\
        \bottomrule
        \end{tabular}}\label{tab:sr}
    \end{center}
    \vspace{-8mm}
\end{table*}

In this context, \( q_t \) is computed based on the predictive uncertainty and prediction confidence at each step.  
The predictive uncertainty is evaluated using the Kullback-Leibler (KL) divergence between the intermediate ensemble predictions at step \( t \) and step \( t-1 \). Specifically, the KL divergence quantifies the change in the probability distributions of the predictions, reflecting how the ensemble output evolves as additional models contribute.  
Mathematically, the predictive uncertainty at step \( t \) can be defined as:
\begin{equation}
    Uncertainty_t = KL(\hat{y}^t_{ens}||\hat{y}^{t-1}_{ens})
\end{equation}
When this value exceeds a certain threshold, it is considered that the prediction is unstable.  
For prediction confidence, we evaluated the confidence of the ensemble prediction at step \( t \) based on a specific threshold. The confidence calculation utilizes the maximum probability of the softmax distribution, which is described as:
\begin{equation}
    Confidence_t = \max(softmax(\alpha^t_c))^2
\end{equation}
Here, \( \text{softmax}(\alpha^t_c) \) represents the class probabilities, and the square of the highest probability indicates the confidence in the prediction.
Ultimately, combining the above elements, \( q_t \) can be defined as:
\begin{equation}
    q_t = \text{sig}(a \cdot Uncertainty_t + b \cdot Confidence_t)
\end{equation}
In this manner, \( q_t \) is dynamically adjusted by incorporating factors such as prediction uncertainty and confidence, ensuring both efficiency and robustness.


\section{Additional Experimental Results}\label{sec:apdx_exp}

\subsection{Evaluation on Larger Datasets}
Tab. \ref{tab:large} compares various AP methods and the proposed EED under a specific sparsity rate \( s_r = 90\% \) condition, evaluated on the CIFAR-100 and ImageNet datasets. In this experiment based on the ResNet50 model, EED demonstrated either superior or comparable performance to existing AP methods across all major metrics. Notably, on the CIFAR-100 dataset, EED achieved a clean accuracy of 63.60\%, showing the highest defense performance against various attacks, including PGD, AutoAttack (AA), and DeepFool. Furthermore, in the case of ImageNet, EED achieved a clean accuracy of 58.41\%, with 30.54\% defense against PGD and 26.89\% defense against AA, demonstrating its outstanding robustness.

Additionally, EED maintained efficient computational speed while offering competitive performance compared to AP methods. EED recorded an average speed-up of 2.90x on CIFAR-100 and 2.84x on ImageNet, showing more efficient utilization of computational resources compared to previous methods. While HARP and TwinRep demonstrated competitive performance, EED outperformed them in terms of both robustness and efficiency. This highlights EED’s ability to deliver high performance and computational efficiency even on large-scale datasets, underscoring its scalability and practical applicability.

\begin{table}[t]
    \begin{center}
    \caption{Evaluation of EED with various AT methods on CIFAR-10 dataset when $s_r = 80\%$}
    \fontsize{8.0pt}{9.0pt}\selectfont
    {\setlength\tabcolsep{4pt} 
    \begin{tabular}{c|c|cccccc}
        \toprule
        \multirow{2}{*}{ } & \multirow{2}{*}{Setting} &\multirow{2}{*}{Clean} &  \multirow{2}{*}{PGD}  &  \multirow{2}{*}{AA}   &  \multirow{2}{*}{C\&W}   &  \multirow{2}{*}{DF}   & Speed \\ &&&&&&&up\\
        \midrule
        \multirow{6}{*}{Resnet18}                                                     
        & Madry\cite{madry2018PGD}               & 87.05 & 56.14 & 48.02 & 57.60 & 53.10 &   1.00x\\
        & EED$_{madry}$                          & 86.13 & 55.71 & 48.13 & 57.03 & 51.97 &   1.86x\\
        \cline{2-8}
        & TRADES\cite{zhang2019theoretically}                          & 85.30 & 57.21 & 51.48 & 52.60 & 51.88 &   1.00x\\
        & EED$_{trades}$                         & 83.84 & 57.41 & 51.53 & 49.65 & 50.02 &   1.84x\\
        \cline{2-8}
        & MART\cite{wang2019improving}                            & 86.16 & 57.72 & 49.39 & 50.53 & 51.59 &   1.00x\\
        & EED$_{mart}$                           & 84.29 & 57.04 & 48.53 & 48.85 & 50.17 &   1.83x\\
        \midrule
        \multirow{6}{*}{VGG16}                                                     
        & Madry\cite{madry2018PGD}               & 82.70 & 54.49 & 48.52 & 54.91 & 56.91 &   1.00x\\
        & EED$_{madry}$                          & 81.39 & 54.26 & 47.49 & 53.27 & 53.94 &   1.79x\\
        \cline{2-8}
        & TRADES\cite{zhang2019theoretically}                          & 83.18 & 55.72 & 49.09 & 54.59 & 57.89 &   1.00x\\
        & EED$_{trades}$                         & 82.13 & 54.79 & 47.26 & 51.98 & 56.30 &   1.80x\\
        \cline{2-8}
        & MART\cite{wang2019improving}                            & 77.44 & 57.51 & 46.20 & 51.38 & 59.56 &   1.00x\\
        & EED$_{mart}$                           & 77.45 & 57.14 & 46.16 & 51.44 & 57.55 &   1.77x\\
        \bottomrule
        \end{tabular}}\label{tab:AT}
    \end{center}
    \vspace{-2mm}
\end{table}

\begin{table}[t]
    \begin{center}
    \caption{Evaluation of EED via ensemble combiner on CIFAR-10 dataset when $s_r = 80\%$}
    \fontsize{8.0pt}{9.0pt}\selectfont
    {\setlength\tabcolsep{4.5pt} 
    \begin{tabular}{c|c|cccccc}
        \toprule
        \multirow{2}{*}{ } & \multirow{2}{*}{Setting} &\multirow{2}{*}{Clean} &  \multirow{2}{*}{PGD}  &  \multirow{2}{*}{AA}   &  \multirow{2}{*}{C\&W}   &  \multirow{2}{*}{DF}   & Speed \\ &&&&&&&up\\
        \midrule
        \multirow{2}{*}{Resnet18}                                                     
        & EED$_{avg}$                            & \textbf{86.13} & \textbf{55.71} & 48.13 & \textbf{57.03} & 51.97 &   1.86x\\
        & EED$_{max}$                            & 85.69 & 55.38 & \textbf{48.52} & 56.27 & \textbf{52.58} &   \textbf{1.88x}\\
        \midrule
        \multirow{2}{*}{VGG16}                                                     
        & EED$_{avg}$                            & \textbf{81.39} & \textbf{54.26} & \textbf{47.49} & 53.27 & 53.94 &   1.79x\\
        & EED$_{max}$                            & 80.83 & 54.11 & 46.70 & \textbf{53.45} & \textbf{55.29} &   \textbf{1.82x}\\
        \bottomrule
        \end{tabular}}\label{tab:combiner}
    \end{center}
    \vspace{-2mm}
\end{table}

\begin{table*}[t]
    \begin{center}
    \caption{Comparison of different ensemble sizes in EED on ResNet18.}
    \fontsize{8.0pt}{9.0pt}\selectfont
    {\setlength\tabcolsep{6 pt} 
    \begin{tabular}{c|c|cccccc|cccccc}
        \toprule
        \multirow{2}{*}{ } & \multirow{2}{*}{Setting} & \multicolumn{6}{c|}{CIFAR-10} &                  \multicolumn{6}{c}{SVHN}\\
                                    & & Clean &  PGD  &  AA   &  C\&W   &  DF   & Speed up  &  Clean &  PGD &  AA   &   C\&W   &   DF   & Speed up \\
        \midrule                                                     
        \multirow{5}{*}{$s_r = 50\%$}
        & $\text{EED}_{d=1}$          & 84.53 & 53.27 & 45.62 & 55.43 & 45.18 & 1.61x & 88.04 & 48.91 & 43.97 & 49.68 & 40.61 & \underline{1.62x} \\
        & $\text{EED}_{d=3}$          & 85.14 & 54.29 & 46.85 & 56.37 & 50.04 & 1.63x & 90.76 & 52.79 & 45.18 & 50.02 & 46.22 & \textbf{1.63x} \\ 
        & $\text{EED}_{d=4}$          & \textbf{88.07} & \underline{57.83} & \underline{52.35} & \textbf{57.92} & \textbf{53.12} & \underline{1.64x} & \textbf{93.15} & \underline{55.74} & \textbf{50.18} & \textbf{58.37} & \underline{56.05} & \textbf{1.63x} \\ 
        & $\text{EED}_{d=5}$          & \underline{87.43} & \textbf{57.87} & \textbf{52.83} & \underline{57.19} & \underline{52.68} & \textbf{1.65x} & 92.71 & \textbf{55.98} & \underline{49.59} & \underline{57.58} & \textbf{56.41} & 1.60x \\ 
        & $\text{EED}_{d=6}$          & 86.24 & 56.21 & 51.89 & 56.12 & 52.81 & 1.59x & \underline{92.81} & 53.21 & 49.14 & 56.53 & 53.32 & 1.57x \\

        \midrule
        \multirow{5}{*}{$s_r = 80\%$}
        & EED$_{d=1}$                 & 81.57 & 53.42 & 45.74 & 55.71 & 45.24 &   1.74x   & 84.18 & 48.97 & 43.89 & 49.82 & 40.73 &  1.72x   \\
        & EED$_{d=3}$                 & 84.22 & 54.11 & 46.59 & 56.22 & 49.88 &   1.78x   & 90.94 & 53.15 & 45.52 & 49.99 & 46.22 &  1.80x   \\
        & EED$_{d=4}$                 & \textbf{86.13} & \textbf{55.71} & \textbf{48.13} & \textbf{57.03} & \textbf{51.97} &   \textbf{1.86x}   & \textbf{93.15} & \textbf{55.74} & \textbf{50.18} & \textbf{58.37} & \textbf{56.05} &  \underline{1.85x}   \\
        & EED$_{d=5}$                 & \underline{84.87} & \underline{55.29} & \underline{47.92} & \underline{56.78} & \underline{51.53} &   1.83x   & \underline{91.83} & \underline{55.12} & \underline{49.87} & \underline{58.89} & \underline{55.43} &  \textbf{1.86x}   \\
        & EED$_{d=6}$                 & 82.44 & 54.03 & 46.72 & 55.95 & 50.29 &   \underline{1.84x}   & 89.75 & 53.34 & 46.38 & 55.12 & 50.89 &  1.83x   \\
        
        \midrule
        \multirow{5}{*}{$s_r = 90\%$}
        & EED$_{d=1}$                 & 79.77 & 48.81 & 42.93 & 51.96 & 45.48 &   2.36x   & 82.02 & 49.18 & 44.19 & 47.85 & 40.72 &  2.31x   \\
        & EED$_{d=3}$                 & 81.33 & 51.28 & \textbf{46.87} & \underline{56.59} & 49.01 &   2.44x   & 90.78 & \underline{53.12} & 45.48 & 51.93 & 48.23 &  2.33x   \\
        & EED$_{d=4}$                 & \textbf{83.26} & \textbf{52.14} & \underline{46.85} & \textbf{56.77} & \textbf{50.21} &   \textbf{2.51x}   & \textbf{91.74} & \textbf{55.16} & \underline{47.32} & \textbf{53.81} & \textbf{53.57} &  \textbf{2.45x}   \\
        & EED$_{d=5}$                 & \underline{82.12} & \underline{51.92} & 46.67 & 56.52 & \underline{49.89} &   \underline{2.47x}   & \underline{91.45} & 52.78 & \textbf{47.41} & \underline{52.28} & \underline{51.04} &  2.40x   \\
        & EED$_{d=6}$                 & 80.95 & 50.18 & 44.32 & 53.48 & 47.77 &   2.46x   & 89.86 & 51.06 & 45.21 & 50.79 & 47.15 &  \underline{2.43x}   \\
        \bottomrule
        \end{tabular}}\label{tab:size}
    \end{center}
    \vspace{-2mm}
\end{table*}

\subsection{Evaluation based on Sparsity}
As discussed in \cref{sec:5.3}, this experiment compares various AP methods with the proposed EED method via sparsity rates and then analyzes the performance on the CIFAR-10 and SVHN datasets using the ResNet18 architecture. The primary comparison factors are the accuracy, robustness against various attacks, and the speed-up ratio, based on the model's sparsity rate (\(s_r\)).

According to Tab. \ref{tab:sr}, across sparsity rates ranging from \(s_r = 50\%\) to \(s_r = 95\%\), EED consistently outperforms other AP methods, with particular emphasis on clean accuracy and defense performance against PGD, AutoAttack, and DeepFool. These results suggest that EED maintains robustness even in sparsely compressed models through efficient and diverse ensemble strategies. At each sparsity rate, EED enhances robustness by utilizing differentiated pruning techniques and assigning diverse data subsets across AP-based sub-models, while improving efficiency through dynamic ensemble selection during the inference stage. Although performance is expected to decrease as \(s_r\) increases, EED maintains relatively high and stable performance compared to competing methods.

Notably, at a lower compression rate, i.e., \(s_r = 50\%\), EED showed overall higher performance than not only other AP methods but also the pre-compressed AT model. For instance, on the CIFAR-10 dataset, EED outperformed AT by 1.69\% in PGD and 4.33\% in AA. Moreover, EED demonstrated significant inference speed-up, with a 1.64x speed-up on CIFAR-10 and a 1.63x speed-up on SVHN compared to other AP methods, reflecting a substantial improvement in computational efficiency. This performance boost is likely due to the impact of the DIE getting stronger as the ensemble size increases.

However, EED's performance decreased as the compression rate increased, primarily due to the reduction in ensemble diversity. Despite this, at \(s_r = 90\%\), EED still outperformed most AP methods, and even at the extreme compression rate of \(s_r = 95\%\), where methods like RST and MAD showed significant performance degradation, EED demonstrated robust defense performance with only minor performance decline.

\subsection{Evaluation based on other AT methods}
According to Tab. \ref{tab:AT}, EED, when combined with existing AT methods such as Madry-AT, TRADES-AT, and MART-AT, maintained similar performance across clean and various attacks (PGD, AA, C\&W, DF), while achieving an average speed-up of 1.77x to 1.86x. Notably, in both ResNet18 and VGG16, EED successfully improved computational efficiency while minimizing performance degradation compared to the base model. For instance, EED$_{MART}$, compared to MART-AT, showed a slight improvement in clean accuracy on VGG16, and defense performance against attacks remains similar, while achieving a 1.77x inference speed-up. This indicates that EED has high compatibility with various AT methods and can be easily integrated with other AT approaches, providing further model compactness and better defense performance.

\subsection{Average Combiner vs Max Combiner}
As mentioned in \cref{sec:4.2}, the final output of EED is computed using an average operation. In this experiment, we compared two ensemble output calculation methods: the average combiner \( h(x) = \frac{1}{N} \sum_{i=1}^{N} h^i(x) \) and the maximum combiner \( h_j(x) = \max_{i \in [N]}(h^i_j(x)) \).

For ResNet18, the average combiner (EED${avg}$) showed relatively better performance in terms of clean accuracy, as well as defense against PGD and C\&W attacks. However, the max combiner (EED${max}$) demonstrated stronger defense against AA and DF attacks. In terms of speed-up, the max combiner achieved a slightly higher improvement of 1.88x. Similar trends were observed in VGG16, where the average combiner provided more stable performance in most cases, but the max combiner showed stronger defense against certain attacks (C\&W and DF).

The results in Tab. \ref{tab:combiner} demonstrate that both combiners can enhance EED's performance while maintaining its compactness and efficiency. This suggests that both the average and max combiners extend the diversity and flexibility of EED, providing different advantages in specific attack scenarios.

\begin{figure*}[t]
  \centering
  \includegraphics[width=1\linewidth]{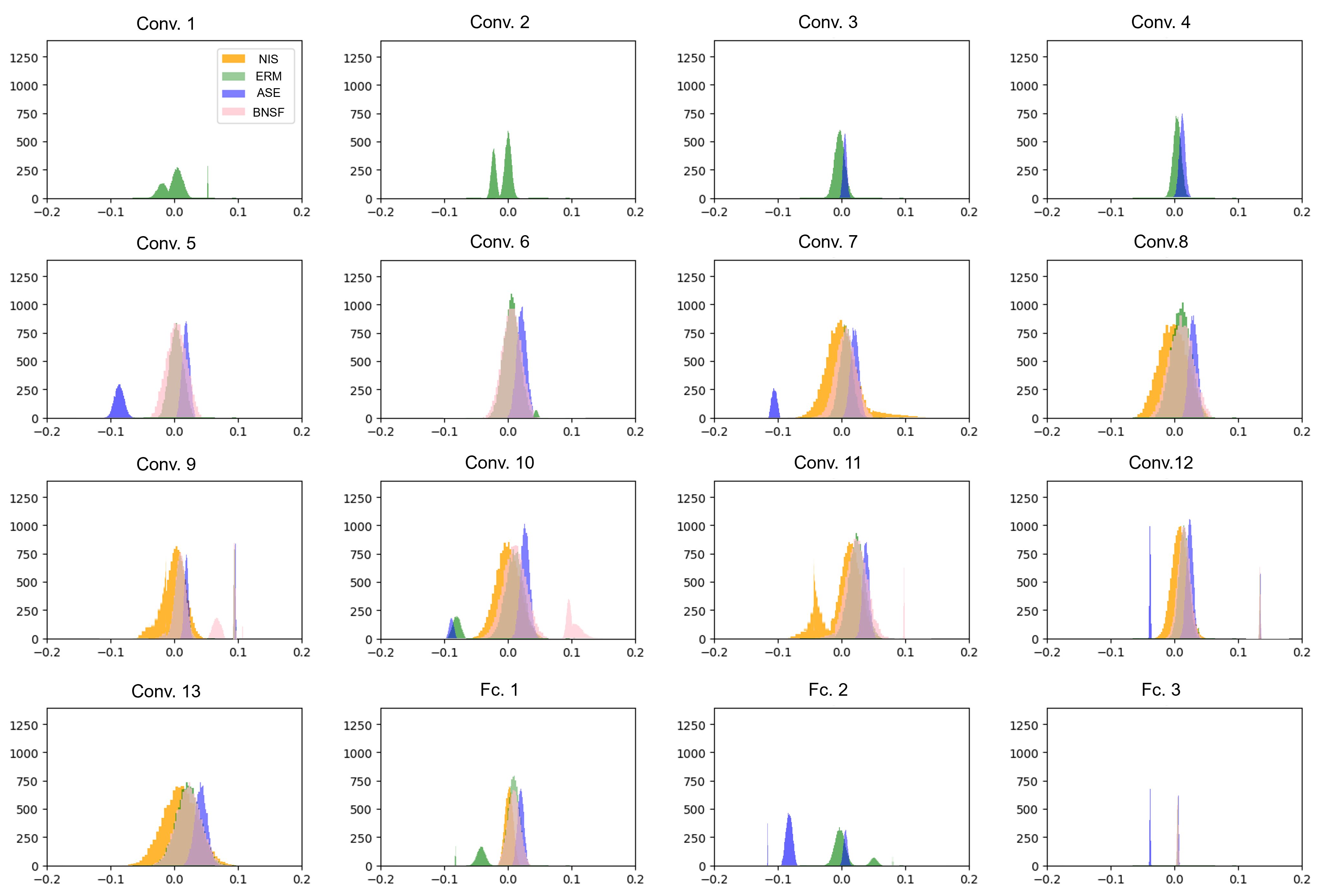}
  \caption{Parameter distribution in each layer of VGG16 pruned to 90\% sparsity with AT.}
  \label{fig:layer}
\end{figure*}

\subsection{Evaluation based on Ensemble Sizes}
As mentioned in \cref{sec:4.1}, we divide the training dataset for pruning into multiple subsets, and assign each subset to train a specific sub-model for the ensemble diversity. In the main paper, we set the number of subsets, \( d \), to 4, and formed ensemble teams with 12 sub-models, denoted as \( |EnsSet| \). Here, we evaluated the impact of the ensemble size on EED by adjusting \( d \).

As shown in Tab. \ref{tab:size}, when \( d=1 \), i.e., when all sub-models are trained on the same dataset and the number of sub-models is small, we observed a general decline in clean accuracy and defense performance against attacks. This indicates that as the number of diverse sub-models in the ensemble decreases, the ensemble's diversity reduces, which in turn limits the ensemble's defensive capability. This performance degradation becomes more pronounced as the sparsity rate increases, which suggests that ensemble diversity becomes increasingly important as the model sparsifies.

On the other hand, as \( d \) increases, model diversity improves, with \( d=4 \) or \( d=5 \) typically yielding the best performance. These configurations strike a balance between sufficient data sharing and the diversity of individual sub-models, which maximizes defense performance against attacks. However, when \( d \) was too large (e.g., \( d=6 \)), the amount of shared data between sub-models decreased, leading to insufficient training data for each sub-model. Consequently, model performance deteriorated, and defense capability weakened. This suggests that overly dividing the dataset to increase the number of sub-models may prevent individual models from learning essential core data, thereby weakening the overall defense capability of the ensemble.

Therefore, selecting an appropriate value for \( d \) plays a critical role in maximizing the defense performance and efficiency of EED, and it is necessary to maintain a balance between data diversity and the amount of shared data.

\subsection{Analysis on Pruning Score via Parameter Distribution}

We used multiple pruning importance scores to promote submodel diversity. In selecting the scores, we focused on their widespread use and effectiveness and their distinctiveness from each other.
Fig.\ref{fig:layer} illustrates the distribution of parameters across different layers of an adversarially trained VGG16 model, which has been pruned to 90\% sparsity on the CIFAR-10 dataset. The figure focuses on analyzing the effects of different pruning importance scores (NIS, ERM, ASE, BNSF). These distribution differences are closely related to the strategy employed in EED, where each importance score is used to maximize the model's diversity.

The parameter distribution varies significantly depending on the pruning importance score, with some criteria concentrating the distribution on specific layers, while others form broader and more diverse distributions. This indicates that each importance score emphasizes different patterns during training, contributing to the model's compressibility and defense efficiency.

These differences help the sub-models generated by applying different pruning scores in EED to achieve greater diversity and robustness. Since each sub-model in EED is trained using a different pruning importance score, the diversity of the overall model pool is ensured. This, in turn, mitigates the vulnerabilities of individual models and enhances the defense effectiveness.

\begin{figure*}[t]
  \centering
  \includegraphics[width=1\linewidth]{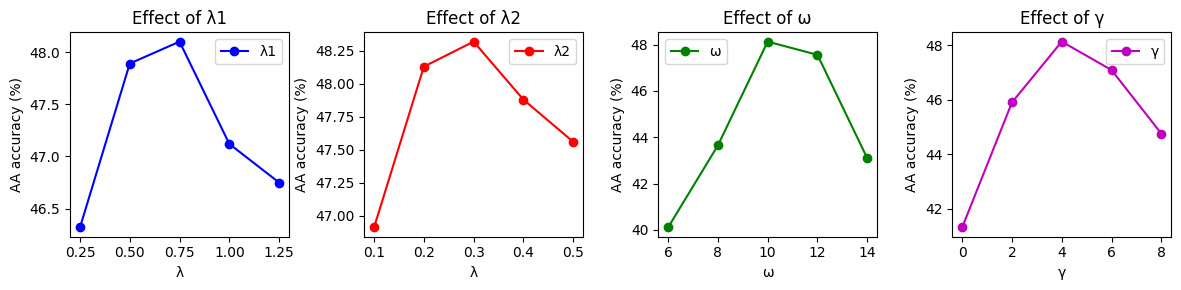}
  \caption{Effect of each hyperparameters in EED.}
  \label{fig:hyper}
\end{figure*}

\subsection{Analysis on Hyperparpameters}
Fig.\ref{fig:hyper} illustrates the effects of various hyperparameters on the accuracy of a classification model. The first graph shows the AA accuracy as a function of $\lambda_1$, reaching its peak at 0.6. The second graph depicts the effect of $\lambda_2$, where optimal performance is observed at 0.3. The third graph presents the accuracy corresponding to changes in $\omega$, with a maximum accuracy achieved at 10. Lastly, the fourth graph illustrating the effect of $\gamma$ shows the highest accuracy at 4. Collectively, these results emphasize the significance of hyperparameter tuning in optimizing model performance, which also indicate the importance of regularization terms.

\section{Other Ensemble Defenses}\label{sec:apdx_ens}

In \cref{sec:4.1}, we have discussed the diversity loss term (\(Div\)) and the regularization term (\(Reg\)) in existing ensemble defense techniques. In this section, we briefly describe four major ensemble-based adversarial defense methods, including ADP \cite{pang2019improving}, GAL \cite{kariyappa2019improving}, DVERGE \cite{yang2020dverge}, and SoE \cite{cui2022synergy}, and evaluate the performance of EED, when applying the \(Div\) and \(Reg\) terms used in these methods, along with the DVERGE technique employed in the main paper.

\subsection{ADP}
ADP employs an ensemble approach through averaging, defined as \( h(x) := \frac{1}{N} \sum_{i=1}^{N} h_i(x) \). The base classifiers are trained to minimize a loss function consisting of the cross-entropy loss for each classifier, a regularization term based on the Shannon entropy of the ensemble prediction, and a diversity loss that encourages various predictions. The loss function for the training example \( (x, y_x) \) is defined as:
\begin{equation}
    \begin{split}
        \mathcal{L}_{\text{ADP}}(x, y_x) = \sum_{i=1}^{N} \ell_{\text{CE}}(h_i(x), y_x) - \alpha Reg(h(x)) \\+ \beta \log Div(h_1(x), h_2(x), \ldots, h_N(x), y_x)
    \end{split}
\end{equation}
Here, the Shannon entropy is \( \text{Reg}(p) = -\sum_{i=1}^{C} p_i \log(p_i) \). The diversity term \( D(h_1, h_2, \ldots, h_N, y) \) measures the geometric diversity between the \( N \) distinct \( C \)-dimensional probability vectors. To compute the diversity, a normalized \( (C-1) \)-dimensional vector \( \tilde{h}_i \) is derived by removing the element at the \( y \)-th position from \( h_i \), and these vectors are stored as columns in the \( (C-1) \times N \) matrix \( \tilde{Reg}_y \). The diversity measure is computed as \( \det(\tilde{Reg}_y^T \tilde{Reg}_y) \).

\subsection{GAL}
GAL proposes a loss function that considers both the prediction diversity and gradient diversity among the base classifiers. The loss function for the training example \( (x, y_x) \) is defined as follows:
\begin{equation}
    \begin{split}
        \mathcal{L}_{\text{GAL}}(x, y_x) =  \frac{1}{N} \sum_{i=1}^{N} D_{\text{KL}}(s_i(x) \| h_i(x)) \\ 
        - \alpha \log \left( \frac{2}{N(N-1)} \sum_{i=1}^{N} \sum_{j=i+1}^{N} \exp\left(JSD(s_i(x) \| s_j(x))\right)\right) \\
        + \frac{2\beta}{N(N-1)} \sum_{i=1}^{N} \sum_{j=i+1}^{N} \cos(\nabla D_{\text{KL}}
        (s_i(x) \| h_i(x)), \\\nabla D_{\text{KL}}(s_j(x) \| h_j(x)))
    \end{split}
\end{equation}
where \( s_i(x) \) is the soft label vector obtained by label smoothing, and \( JSD \) is the Jensen-Shannon divergence.

\subsection{DVERGE}
DVERGE emphasizes the vulnerability diversity between the base classifiers to provide better adversarial robustness. For the \(i\)-th base classifier, the following is minimized:
\begin{equation}
    \begin{split}
        \mathcal{L}_{\text{DVERGE}}(x, y_x) = \ell_{\text{CE}}(h_i(x), y_x) \\+ \alpha \sum_{j \neq i} \mathbb{E}_{(x_s, y_{x_s}) \sim D, l \in [L]} \left[\ell_{\text{CE}}(h_i(\tilde{x}_{h_j(l)}(x, x_s)), y_{x_s})\right]
    \end{split}
\end{equation}
Here, $\tilde{x}_{h_j(l)}$ reflects non-robust features.

\subsection{SoE}
SoE is based on vulnerability diversity and two training phases utilizing different adversarial examples. The loss function for the \(i\)-th base classifier \(h_i\) is defined as:
\begin{equation}
    \begin{split}
        \mathcal{L}_{\text{SoE}}(x, y_x) = \sum_{j=1}^{N} \ell_{\text{BCE}}(h_j(y_x, \tilde{x}_i), g_j(\tilde{x}_i)) \\- \sigma \ln \sum_{j=1}^{N} \exp\left(-\frac{\ell_{\text{CE}}(h_j(\tilde{x}_i), y_x)}{\sigma}\right)
    \end{split}
\end{equation}

\subsection{Analysis on Ensemble Defenses}
\begin{table}[t]
    \begin{center}
    \caption{Evaluation of EED with various ED losses on CIFAR-10 dataset when $s_r = 80\%$}
    \fontsize{8.0pt}{9.0pt}\selectfont
    {\setlength\tabcolsep{4.5pt} 
    \begin{tabular}{c|c|cccccc}
        \toprule
        \multirow{2}{*}{ } & \multirow{2}{*}{Setting} &\multirow{2}{*}{Clean} &  \multirow{2}{*}{PGD}  &  \multirow{2}{*}{AA}   &  \multirow{2}{*}{C\&W}   &  \multirow{2}{*}{DF}   & Speed \\ &&&&&&&up\\
        \midrule
        \multirow{4}{*}{Resnet18}
        & EED$_{\text{ADP}}$                            & 84.90 & 54.23 & 46.87 & 55.12 & 50.20 &   1.86x\\
        & EED$_{\text{GAL}}$                            & 85.30 & 54.70 & 47.50 & 55.75 & 50.90 &   1.85x\\
        & EED$_{\text{DVERGE}}$                         & 86.13 & 55.71 & 48.13 & 57.03 & 51.97 &   1.86x\\
        & EED$_{\text{SoE}}$                            & 85.85 & 55.10 & 47.90 & 56.40 & 52.10 &   1.82x\\
        \midrule
        \multirow{4}{*}{VGG16}
        & EED$_{\text{ADP}}$                            & 80.92 & 53.13 & 45.83 & 52.07 & 52.29 &   1.78x\\
        & EED$_{\text{GAL}}$                            & 81.05 & 53.79 & 46.38 & 52.52 & 53.07 &   1.80x\\
        & EED$_{\text{DVERGE}}$                         & 81.39 & 54.26 & 47.49 & 53.27 & 53.94 &   1.79x\\
        & EED$_{\text{SoE}}$                            & 80.59 & 53.95 & 46.10 & 53.11 & 54.58 &   1.81x\\
        \bottomrule
        \end{tabular}}\label{tab:ensemble}
    \end{center}
    \vspace{-2mm}
\end{table}
We evaluated EED with various ensemble loss functions on the CIFAR-10 dataset. The results shown in Tab.\ref{tab:ensemble} demonstrate that EED has strong compatibility with a range of ensemble defense strategies, delivering notable improvements in adversarial robustness and inference speed. EED consistently showed excellent performance against attack types such as PGD, AA, C\&W, and DF, with particularly high robustness observed when using vulnerability diversity terms like those in DVERGE and SoE, which performed well across all attack types.

\section{Limitations and Future Work}\label{sec:apdx_lim}

The proposed Efficient Ensemble Defense (EED) enhances the diversity of compressed models, maintains robustness against adversarial attacks, and effectively reduces model capacity. However, EED has a few limitations yet.

First, the training cost is relatively high. The process of generating sub-models using multiple pruning metrics and combining them into an ensemble can increase training time compared to conventional Adversarial Pruning (AP). This is an inevitable aspect of ensemble defense, but when compared to ensembles using multiple base models, the cost is significantly lower. Unlike AP, which requires a lot of cost for fine-tuning, EED can address this by leveraging the ensemble, thus mitigating the overhead during sub-model generation. However, when using large-scale datasets or complex models, it can still be difficult to train models in environments with limited computing resources.

Additionally, when applying extreme sparsity, as shown in \cref{sec:5.3} and Tab.\ref{tab:sr}, there is a tendency for model performance to degrade. Although EED aims to use compressed models for efficiency, applying excessively high sparsity can lead to performance degradation. This happens because, as sparsity increases, ensemble diversity decreases, which diminishes the advantages of the ensemble approach.

Future research could explore several approaches to overcome these limitations. First, optimization methods for reducing training costs should be considered. In particular, efficient algorithms are required to reduce the computational overhead in the sub-model creation and ensemble combination process. For example, techniques that optimize parameter sharing during training or streamline the sub-model selection process could help shorten training times. Next, more refined pruning techniques need to be developed to address the performance degradation caused by high sparsity. Current pruning methods may damage the model's structure and important parameters as sparsity increases, so exploring ways to maximize compression while minimizing performance loss is crucial. Additionally, techniques that maintain diversity among sub-models, even with high compression, could be developed.

\end{document}